\documentclass[preprint,authoryear,a4paper]{elsarticle}

\usepackage{natbib}

\usepackage{verbatim}

\usepackage{geometry}

\usepackage{color,soul}
\usepackage{url}

\usepackage[hidelinks]{hyperref}
\hypersetup{
  colorlinks   = true, 
  urlcolor     = blue,
  linkcolor    = blue, 
  citecolor    = blue 
}

\usepackage{graphicx}
\graphicspath{{./images/}}
\DeclareGraphicsExtensions{.pdf,.png}
\usepackage{placeins}
\usepackage{caption}
\usepackage{subfigure}
\usepackage{minipage-marginpar}
\usepackage{booktabs}
\usepackage[table,xcdraw]{xcolor}
\usepackage{graphicx}
\usepackage{adjustbox}
\usepackage{multirow}

\usepackage{amsbsy}
\usepackage{amssymb}
\usepackage{gensymb}
\usepackage[mathscr]{euscript}
\usepackage{bm}
\usepackage{mathtools}
\usepackage{nccmath}
\usepackage{amsmath}
\usepackage{braket}

\usepackage{algorithm}
\usepackage{algpseudocode}

\usepackage{lineno,hyperref}

\usepackage{lipsum}

\journal{ISPRS Journal of Photogrammetry and Remote Sensing}

\begin{document}

\begin{frontmatter}

  \title{Leveraging Photogrammetric Mesh Models for Aerial-Ground Feature Point Matching Toward Integrated 3D Reconstruction}

  \author[swjtu]{Qing Zhu}
  \author[swjtu]{Zhendong Wang}
  \author[swjtu]{Han Hu\corref{cor1}}
  \cortext[cor1]{Corresponding Author: han.hu@swjtu.edu.cn}
  \author[szu]{Linfu Xie}
  \author[swjtu]{Xuming Ge}
  \author[whu]{Yeting Zhang}

  \address[swjtu]{Faculty of Geosciences and Environmental Engineering, Southwest Jiaotong University, Chengdu, China}
  \address[szu]{Guangdong Key Laboratory of Urban Informatics \& Shenzhen Key Laboratory of Spatial Smart Sensing and Services \& Research Institute for Smart Cities, School of Architecture and Urban Planning, Shenzhen University, Shenzhen, China}
  \address[whu]{State Key Laboratory of Information Engineering in Surveying, Mapping and Remote Sensing, Wuhan University, Wuhan, China}

  \begin{abstract}
    Integration of aerial and ground images has been proved as an efficient approach to enhance the surface reconstruction in urban environments. However, as the first step, the feature point matching between aerial and ground images is remarkably difficult, due to the large differences in viewpoint and illumination conditions. Previous studies based on geometry-aware image rectification have alleviated this problem, but the performance and convenience of this strategy are still limited by several flaws, \emph{e.g.} quadratic image pairs, segregated extraction of descriptors and occlusions. To address these problems, we propose a novel approach: leveraging photogrammetric mesh models for aerial-ground image matching. The methods have linear time complexity with regard to the number of images. It explicitly handles low overlap using multi-view images. The proposed methods can be directly injected into off-the-shelf structure-from-motion (SFM) and multi-view stereo (MVS) solutions. First, aerial and ground images are reconstructed separately and initially co-registered through weak georeferencing data. Second, aerial models are rendered to the initial ground views, in which color, depth and normal images are obtained. Then, feature matching between synthesized and ground images are conducted through descriptor searching and geometry-constrained outlier removal. Finally, oriented 3D patches are formulated using the synthesized depth and normal images and the correspondences are propagated to the aerial views through patch-based matching. Experimental evaluations using five datasets reveal satisfactory performance of the proposed methods in aerial-ground image matching, which succeeds in all of the ten challenging pairs compared to only three for the second best. In addition, incorporation of existing SFM and MVS solutions enables more complete reconstruction results, with better internal stability.
  \end{abstract}

\end{frontmatter}

\section{Introduction}
\label{s:intro}

\begin{figure}[ht]
  \centering
  \includegraphics[width=\linewidth]{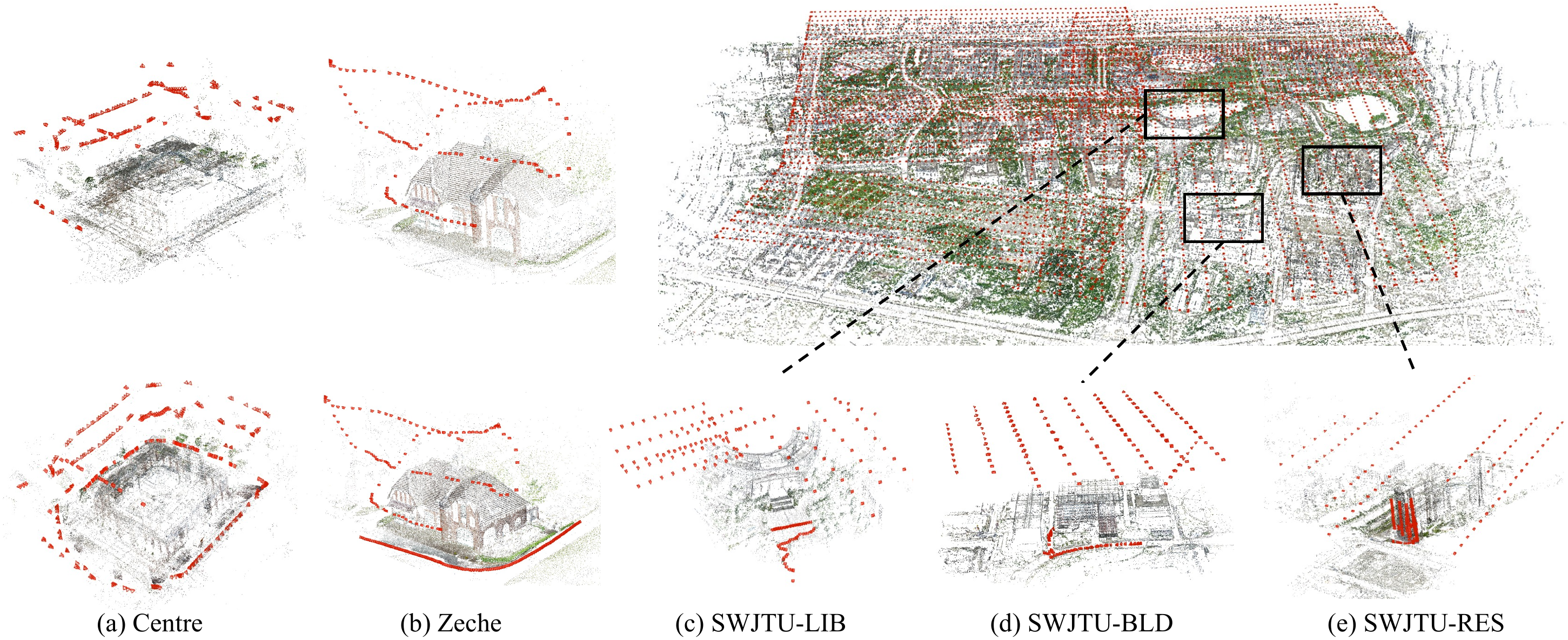}
  \caption{Aerial-ground reconstruction for the ISPRS benchmark \citep{nex2015isprs} and three buildings of the Southwest Jiaotong University (SWJTU), Chengdu, China. The top row depicts the different structures of aerial image collections and the bottom row shows the reconstructed aerial and ground images. The images are rendered using Colmap \citep{schoenberger2016sfm}.}
  \label{fig:acquisition}
\end{figure}

Penta-view aerial oblique images \citep{lemmens2014oblique} have become a major source of data for city-scale urban reconstruction. However, occlusion and viewpoint differences greatly perturb the bottom parts of buildings, leading to holes in geometry and texture-blurring effects \citep{wu2018integration}. Recent studies \citep{nex2015isprs,wu2018integration,gao2018ancient} have confirmed that integration of aerial and ground images is a promising approach toward improved 3D reconstruction (see Figure \ref{fig:acquisition}).

The major obstacle to aerial-ground integration is the large viewpoint difference between the two sets of images. It is difficult to find enough tie-points to register both datasets into the same coordinate frame in a combined bundle adjustment. Scale invariant feature transform (SIFT) and SIFT-like features \citep{lowe2004sift,arandjelovic2012three,bursuc2015kernel} are incapable of handling large perspective differences \citep{mikolajczyk2005compare}, and learned features \citep{mishchuk2017working,revaud2019r2d2,dusmanu2019d2} cannot greatly extend the classical approach \citep{arandjelovic2012three,schonberger2017comparative}. Although some researchers have pioneered investigations in this area \citep{wu2018integration,gao2018ancient}, we argue that some key problems remain unfulfilled.

\emph{1) Quadratically increased image rectifications.} Warping all of the images to ground \citep{hu2015reliable} is a valid solution for the nadir and oblique views of aerial images, and the feature extraction has an $ O(n) $ complexity with respect to the number of images. However, the ground structure is not applicable in aerial-ground integration. Pairwise rectification is used to remedy this problem \citep{wu2018integration}, by the adoption of virtual fa\c{c}ades. But pairwise rectification leads to a feature extraction of $ O(n^2) $, which is prohibitively high in practice. Furthermore, such fa\c{c}ade structures may be untenable in certain scenarios.

\emph{2) Problem of pairwise rectification}. Even if the aerial and ground images are rectified successfully, feature matching between them still remains a non-trivial task. For pairwise rectification, contents from only two images are involved, which will lead to some problems in feature matching. For instance, the overlapping region may be only a small part of the whole image, and this region may still be affected by occlusion, as seen in the work by \cite{wu2018integration}.

\emph{3) Mode of the data acquisition.} An effective strategy to avoid the problem of aerial-ground feature matching is to systematically design the image acquisition for both datasets \citep{molina2017first}. For instance, collecting images with acceptable convergent angles around the objects of interest is tenable for certain objects, such as the Centre and Zeche datasets \citep{nex2015isprs} in Figure \ref{fig:acquisition}. However, in practice, flights with regular strips are preferred even for regional applications, such as the campus of SWJTU in Figure \ref{fig:acquisition}. Terrestrial images are only captured to improve the quality of objects of interest. Therefore, the perspective deformation between aerial and ground images is inevitable.

In this paper, we leverage the photogrammetric meshes obtained from aerial images to solve the above problems. Accordingly, instead of rectifying the images pairwisely, we directly render the textured meshes onto a virtual camera determined by the ground images. The rendered images also consist of depth values and normal vectors, and act as proxies between the ground and aerial images. Feature matches are conducted between the ground and rendered images. The correspondences are then enriched with depth and normal information, which can formulate 3D patches in the object space. The 3D patches are then propagated to the aerial images via multi-photo geometrically constrained (MPGC) matching \citep{zhang2005automatic} or patch-based matching \citep{furukawa2009accurate}. A single rendered image contains textural information from multiple aerial images, which are typically selected meticulously in the multi-view stereo (MVS) pipeline \citep{vu2011high,waechter2014let}; therefore, the proposed methods are explicitly occlusion-aware. Additional features are detected only from the rendered images and descriptor matchings are conducted only on the pairs of rendered and ground images; therefore, both feature extraction and feature matching have time complexity of $ O(n) $, with respect to the number of ground images. To handle the illumination differences that lead to degraded descriptor performances, we add an additional filter prior to random sample consensus (RANSAC) \citep{moisan2012automatic} using geometry constraints.

In summary, our main contribution is a simple, fast, accurate and robust approach that solves the problem of aerial-ground feature point matching by rendering the textured mesh models. The reminder of this paper is organized as follows. In Section 2 we briefly describe feature point matching between aerial and ground images. In Section 3 we elaborate on the two steps of the proposed methods, \emph{i.e.} rendering and matching. Experimental evaluations for both the ISPRS datasets \citep{nex2015isprs} and SWJTU datasets are demonstrated (Figure \ref{fig:acquisition}) in Section 4. Finally, concluding remarks are given.

\section{Related works}

Here, we review only directly relevant studies on feature point matching methods in the context of large perspective differences. Specifically, three major strategies for image matching are considered, namely: 1) affine invariant features; 2) image rectification; and 3) 3D rendering. More detailed reviews and comparisons can be found in recent benchmark works \citep{schonberger2017comparative}.

\paragraph{1) Affine invariant features} Following the route of scale and rotation invariant SIFT features \citep{lowe2004sift}, earlier researchers sought affine invariant regions to alleviate perspective deformations. Affine invariant features are generally represented as ellipses on the image \citep{mikolajczyk2004scale,matas2004robust,ma2015affine}. These affine invariant regions may also be detected by line structures \citep{chen2013robust}.  However, in practice, affine invariant detectors are more sensitive to image noise and their repeatability is inferior to that of the difference of Gaussian (DoG) detectors \citep{lowe2004sift} or other corner detectors \citep{rublee2011orb,rosten2010faster}. Therefore, the overall performances of affine invariant detectors are generally worse than those based on SIFT-like features \citep{lowe2004sift}.

\paragraph{2) Image rectification} When no \emph{a priori} geometry information is available, affine SIFT (ASIFT) \citep{morel2009asift} can be used to create a database of descriptors by synthesizing the image in a series of pre-defined affine transformations. A similar approach is used in the database BRIEF \citep{calonder2012brief}, which retrieves BRIEF features on multiple scales and orientations. \cite{roth2017wide} also synthesized a series of views using pairwise perspective transformation and the features are detected using similar sampling strategies as ASIFT \citep{morel2009asift}. However, ASIFT will significantly increase the number of features and therefore increase the search space, leading to longer runtimes and lower recall rate.

In most of standard photogrammetric applications, we have access to the initial image poses, from either the global navigation satellite system (GNSS) or from coarse registrations \citep{wu2018integration,gao2018ancient}. The \emph{a priori} geometry information can help us to rectify the images. For aerial oblique images obtained with regular flight strips, we can identify a \emph{view-independent} structure for the rectification, \emph{i.e.} the ground. For \emph{view-independent} rectification, the base plane for all the images is the same and the perspective deformation between the nadir and oblique views can be alleviated by projecting all the images onto the base plane \citep{hu2015reliable}. This strategy is also applicable to unmanned aerial vehicle (UAV) images \citep{jiang2017onboard} or panoramas captured by mobile mapping systems \citep{jende2018fully,javanmardi2017towards}.

\emph{View-independent} rectifications \citep{hu2015reliable,jiang2017onboard} are convenient, as feature extractions and matchings have the same time complexity $ O(n) $, with respect to the original number of images. However, it is not always possible to find a suitable base plane that all the images can be projected to. Therefore, \emph{view-dependent} rectifications \citep{wu2018integration,gao2018ancient} have been proposed to remedy this problem, for which the surface for rectification is determined pair-wisely rather than unified for all the images. \cite{wu2018integration} found virtual fa\c{c}ade structures by fitting planes from the points inside the frustum of camera, and rectified images by projecting both the aerial and ground images onto the fa\c{c}ade planes. The fa\c{c}ade structures are also used by \cite{fanta2019co} for the co-registration of mobile mapping images and aerial oblique images. In addition, 3D structures can also be considered for pairwise rectification. \cite{gao2018ancient} projected ground images onto aerial views, using the triangular meshes as proxies. A similar strategy was also implemented using dense point clouds \citep{shan2014accurate}, by formulating a depth map corresponding to the ground image and warping the image to aerial view in a pixelwise fashion.

However, \emph{view-dependent} rectification also implies that the descriptor must be extracted on the rectified images (which has quadratic time complexity), and also requires computation of the pairwise image rectifications. Such an onerous process is acceptable only for correlation-based feature matching in local windows rather than the whole image. For instance, previous works have rectified local patches to refine the positions of known tie-points or expand them to neighboring regions, such as \emph{e.g.} multi-photo geometrically constrained (MPGC) correlation \citep{zhang2005automatic} and patch-based multi-view stereo (PMVS) \citep{furukawa2009accurate,wu2018integration}.

\paragraph{3) 3D rendering} The above matching methods only use data from a pair of images, regardless of the methods used for image rectification. In the case of aerial-ground integration, the overlapping region of two images may be quite narrow, limiting the recall rate of the descriptor searching. As an alternative, rendering 3D data onto the target view can explicitly utilize information from multiple images and also exploit the massively parallel graphics computing unit (GPU) for efficient implementation. In this context, \cite{untzelmann2013scalable} rendered the sparse point clouds from SIFT matches using the splat representation \citep{sibbing2013sift,gao2018ancient}. However, the sparse point clouds from SFM are not ideal sources for such rendering. 

Recent solutions \citep{acut3d2019context,agisoft2019metashape,schoenberger2016mvs} can generate high resolution textured mesh models, which can be used as better proxies for the feature matching. And learned MVS approaches \citep{yu2020fast,yao2019recurrent} have demonstrated impressive performances on benchmark tests, which are promising alternatives for off-the-shelf MVS solutions. Except for rendered color images, this paper shows that depth and normal information of the meshes can also be preserved during rendering, which further supports the correlation-based local refinement of matches \citep{zhang2005automatic,furukawa2009accurate}.

\section{Aerial-ground feature point matching by leveraging photogrammetric models}

\subsection{Overview of the approach}

Integrated reconstruction from both aerial and ground images relies on the premise that the intrinsic and extrinsic orientation parameters are consistent in the same coordinate frame, which is achieved by a combined bundle adjustment. The foundation of a successful bundle adjustment is accurate and robust matching of tie-points, which faces the problem of large perspective deformation between aerial and ground images. In previous works \citep{wu2018integration,gao2018ancient}, pairwise image rectifications have partially alleviated this problem, for the estimation of rigid transformations. However, due to the amount and quality of inter-platform tiepoints, previous works need \textit{ad hoc} strategies in the SFM and MVS pipeline. For instance, \cite{gao2018ancient} degraded SFM to a rigid transformation and simplified the MVS as fusion of point clouds from different platforms. \cite{wu2018integration} co-registered images from different platforms by weighted bundle adjustment with parameters regularized by the rigid transformation and also only fused point clouds without a full MVS pipeline. In fact, the key problem still remained to be fulfilled, \textit{i.e.} finding enough inter-platform tiepoints for both the SFM and MVS pipelines.

\begin{figure}[h]
  \centering
  \includegraphics[width=\linewidth]{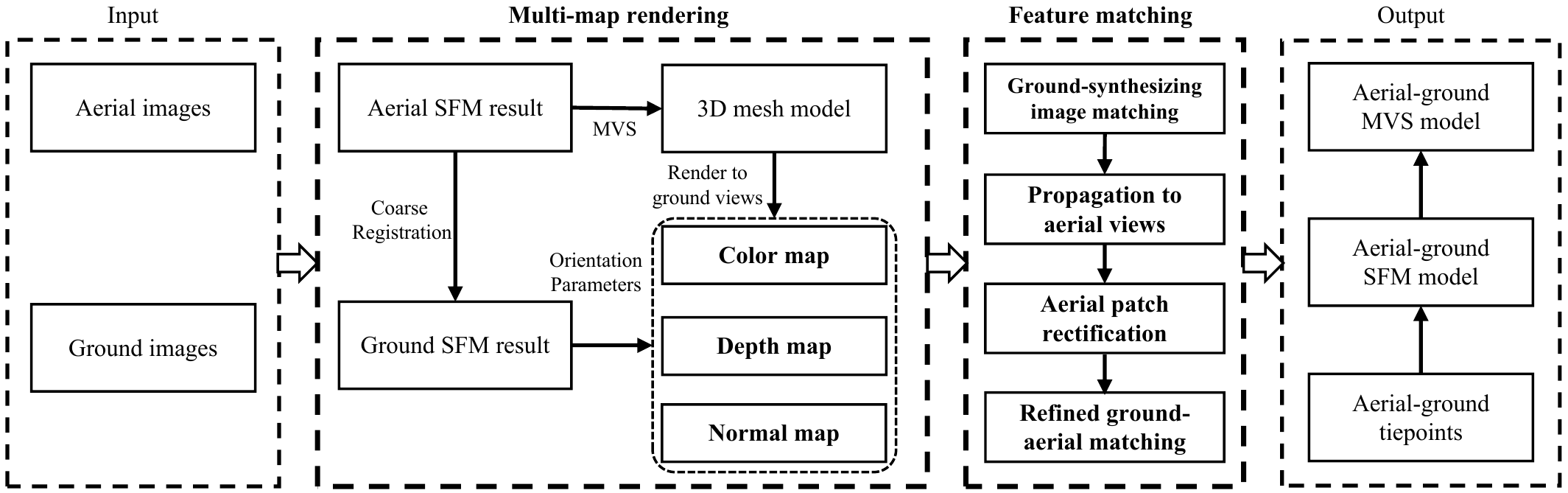}
  \caption{Workflow of the proposed method.}
  \label{fig:overview}
\end{figure}

In this paper we surmount the problem of view-dependent rectification using textured meshes. We render textured meshes to ground images, and use these rendered images as delegates to establish feature matching between aerial and ground images. Figure \ref{fig:overview} demonstrates the overall workflow of the proposed methods. Beginning with two separate datasets, we first reconstruct the sparse models via existing SFM pipeline. Coarse registration is conducted to fuse both aerial and ground models into the same coordinate frame, similar to previous works \citep{wu2018integration,gao2018ancient}; the coarse registration can be achieved by either weak GNSS information or three interactively selected points. As our approach requires no planar structures \citep{wu2018integration}, dense reconstruction using existing MVS pipeline is only required for the aerial datasets, from which tile-wise models are obtained. The textured meshes are rendered using the camera defined by the ground images; the results consist of color, depth and normal vectors. The synthesized color images are matched with the ground images, and correspondences are then propagated to the aerial views using the depth information. Due to insufficient geometric accuracy of the meshes and blending problems of the texture \citep{waechter2014let} in the MVS pipeline, the correspondences have to be refined on the original aerial images. The refinement is achieved through the 3D local patches determined by the depth and normal vectors of the synthesized images. Finally, the matches are directly injected into off-the-shelf SFM and MVS pipelines for integrated reconstruction. 

\subsection{View synthesizing the ground images by rendering of meshes}
\label{s:rendering}

\subsubsection{Definition of the camera models}

To exploit OpenGL graphics pipeline for the synthesis of ground images from textural information of aerial meshes, the notations of intrinsic and extrinsic orientation parameters from SFM and camera matrices of graphics pipeline must be converted.

Specifically, for camera model, we use the protocol of BlockExchange \citep{context2019camera}, in which a 3D point $ \boldsymbol{X} $ is projected to image $ \boldsymbol{x} $ as,
\begin{equation}
  \boldsymbol{x}=fD(\Pi(\textbf{R}(\boldsymbol{X}-\boldsymbol{C})))+\boldsymbol{x}_0,
\end{equation}
where $ f $ and $ \boldsymbol{x}_0 $ are the principal distance and principal point measured in pixels, respectively; $ D(\cdot) $ is the distortion mapping from an undistorted focal plane coordinate to the distorted position and the Brown model with five parameters $ (k_1,k_2,k_3,p_1,p_2) $ is considered; $ \Pi(\cdot): \mathbb{R}^3 \mapsto \mathbb{R}^2 $ is the projection function mapping the 3D point in camera space to the homogeneous normalized coordinate; and $ \textbf{R} $ and $ \boldsymbol{C} $ denote the extrinsic orientation for the rotation matrix and projection center, respectively. In addition, each image is enriched by three depth values recorded in the BlockExchange format, in terms of the nearest $ z_n $, furthest $ z_f $ and median $ z_m $ depth; even without these values, it is trivial to estimate the depth information from the sparse point clouds or the bounding box of the region of interest.

\subsubsection{Estimation of the rendering matrices for the view synthesis}

In the graphics pipeline, the homogeneous coordinate $ \tilde{\boldsymbol{X}} \in \mathbb{R}^4 $ of the 3D point $ \boldsymbol{X} $ is projected to the normalized screen space $ \boldsymbol{m} \in \mathbb{R}^3 $ (and the homogeneous coordinate $ \tilde{\boldsymbol{m}} \in \mathbb{R}^4 $) using view $ \textbf{V} \in \mathbb{R}^{4\times4} $ and projection $ \textbf{P} \in \mathbb{R}^{4\times4} $ matrices as below:
\begin{equation}
  \tilde{\boldsymbol{m}} = \textbf{P}\textbf{V}\tilde{\boldsymbol{X}},
  \label{eq:ogl_projection}
\end{equation}
where the view matrix $ \textbf{V} $ is defined with three parameters, \textit{i.e.} eye $\boldsymbol{E}$, center $ \boldsymbol{O} $ and up $ \boldsymbol{U} $ vectors. The matrix is generally implemented in the \emph{lookat} routine \citep{glm2019opengl}, which describes the position and orientation of the camera. The projection matrix $ \textbf{P} $ is defined by the \emph{perspective} routine \citep{glm2019opengl} using the field of view $ \theta $, aspect ratio $ \rho $, nearest $z_n$ and furthest $z_f$ depth values, which describes the frustum of the camera.  Although it is possible to consider the principal point offsets and distortion of the camera in the graphics pipeline by exploiting the program shaders, we ignore them for two reasons: (1) the influences of them on perspective deformation are almost negligible and (2) they only influence the intermediate coordinates on the synthesized images, which will be eventually propagated to aerial views and refined.

To obtain the eye $\boldsymbol{E}$, center $ \boldsymbol{O} $ and up $ \boldsymbol{U} $ vectors for the \emph{lookat} function, the conversion is determined intuitively as:
\begin{equation}
  \begin{split}
    \boldsymbol{E}&=\boldsymbol{C}\\
    \boldsymbol{O}&=\boldsymbol{C} + z_m\textbf{R}^T\boldsymbol{e}_z\\
    \boldsymbol{U}&=-\textbf{R}^T\boldsymbol{e}_y
  \end{split},
\end{equation}
where $ \boldsymbol{e} $ denotes the unit vector along the corresponding axis and $ \textbf{R}^T $ transforms the axis in camera coordinate space to object coordinate space. With respect to the parameters in the \emph{perspective} function, $ z_n $ and $ z_f $ are directly used for the depth range and the other two parameters are calculated as:
\begin{equation}
  \begin{split}
    \theta&= 2\arctan{\frac{h}{2f}}\\
    \rho&=\frac{w}{h}
  \end{split},
\end{equation}
where $ w $ and $ h $ are the width and height of the images, respectively.

\subsubsection{Rendering of the color, depth and normal images}

\begin{figure}[htbp]
    \centering
    \includegraphics[width=\linewidth]{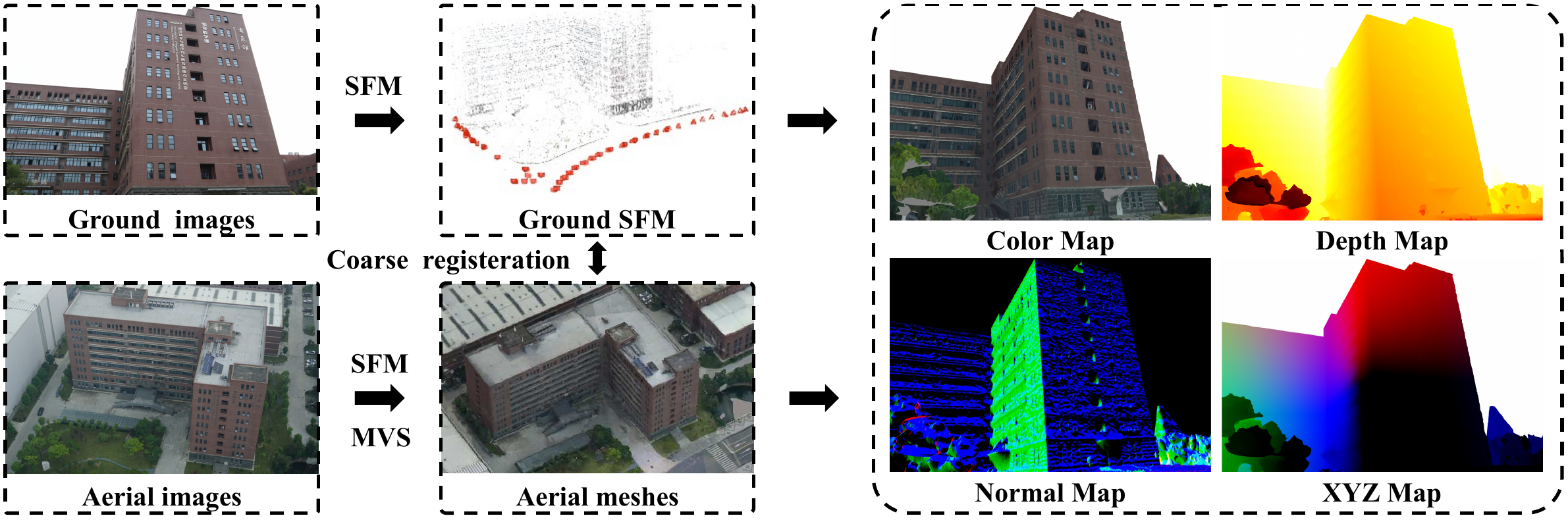}
    \caption{Illustration of the rendering of the meshes to various maps, comprising color images, depth images and normal images. The coordinates of each pixel in the rendered image can be obtained as the XYZ map.}
    \label{fig:render}
\end{figure}

Another practical issue for the rendering of the textured meshes is that the meshes are tiled on a tree structure, \textit{e.g.} quad-tree, octree or adaptive KD-tree. Even inside a single tile, the models are still segmented into small pieces with different level-of-details to accelerate the loading of files from disks. The render engine should use a scene graph to organize the dynamic loading (or unloading) of the meshes that are inside (or outside, respectively) the frustum of current view. This is non-trivial in implementation, but fortunately, OpenSceneGraph \citep{osfield2004open} has already implemented an optimized database manager with its native data format. For each frame, we wait for the database manager to fully load the load the finest level of detail of model in the current view, before actually rendering the models. For the rendering, we allocate three frame-buffer objects to store the color, depth and normal information (Figure \ref{fig:render}), and the meshes are then directly rendered to the buffers rather than to the physical screen. The sizes of the frame-buffer objects are the same as those of the corresponding cameras, therefore reducing the differences of scale and other geometric factors.

Notably, the rendering of the meshes explicitly utilizes the massively parallel GPU and can be achieved almost in real time. In addition, any point in the color image is one-on-one mapped to the 3D object space with the depth map (XYZ map in Figure \ref{fig:render}). Therefore, by enriching a point with a normal vector, we can obtain a locally oriented 3D patch; this is similar to the concept of previous work \citep{furukawa2009accurate}. The patch is helpful for the refinement of correspondences between aerial and ground images.

\subsection{Feature matching and refinement with the synthesized images}
\label{s:matching}

\begin{figure}[p]
	\centering
	\includegraphics[width=\linewidth]{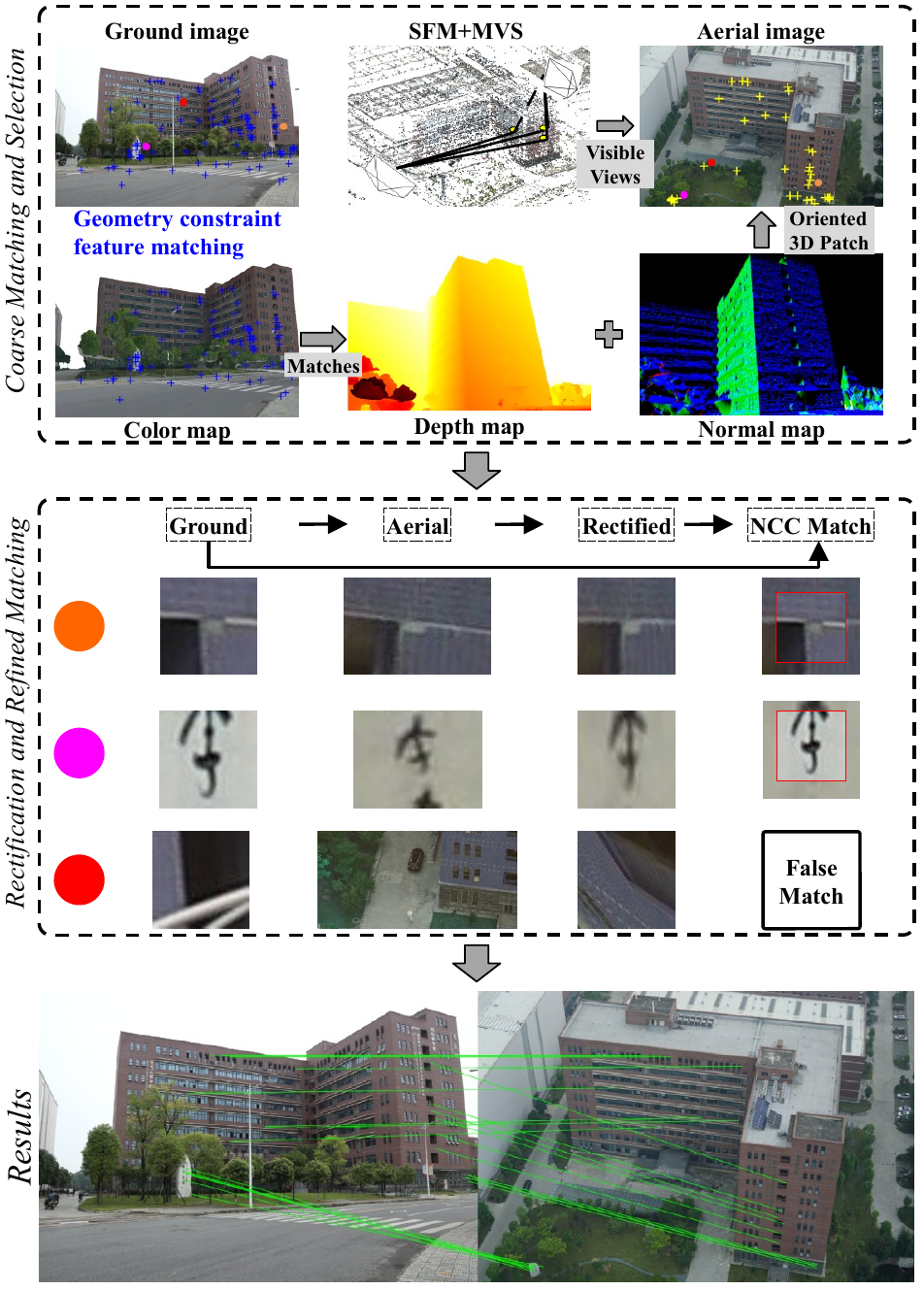}
	\caption{Overview of aerial-ground feature matching. The circles in the coarse-matching images denote the three patches in the refined matching.}
	\label{fig:match_workflow}
\end{figure}

Figure \ref{fig:match_workflow} illustrates the two steps of the aerial-ground feature-point matching. For coarse matching, we first extract SIFT features \citep{lowe2004sift} on the synthesized images, because SIFT is still the default option in many solutions \citep{wu2011visualsfm,schoenberger2016sfm}. Then, we compare descriptors between the ground and synthesized images, using the ratio check and filter outliers, using both the proposed geometrical constraints (Subsection \ref{s:greedy_outlier}) and RANSAC \citep{fischler1981random}. Specifically, we use a recent variant of RANSAC, the \emph{a contrario} RANSAC (AC-RANSAC), which features automatic threshold tuning \citep{moisan2012automatic}. If the remaining number of pairwise matches between the synthesized and ground images is less than five, we consider the matching to be not stable and ignore the results for this pair. 

3D patches are formulated using the depth and normal information from matches on the synthesized images. The coordinate $ \boldsymbol{X} $ in 3D space is calculated from the corresponding depth value using the reverse of Equation \ref{eq:ogl_projection}. The ground sample distance $ \delta=\frac{d}{f}$ is also estimated from the depth value $d$. We assign a relatively large search window $ w_s\delta $ in the object space as delegates, which is centered on and tangential to the oriented points $ (\boldsymbol{X},\boldsymbol{n}) $. In the following section, we use the term $ p=(\boldsymbol{X},\boldsymbol{n},w_s\delta) $ to denote the oriented patches in the object space, inspired by previous work \citep{furukawa2009accurate}. Suitable views of the aerial images are selected (Subsection \ref{s:propagate}) for each local patch and then the patch is projected to aerial views for subsequent refinement.

For refined matching (Subsection \ref{s:aerial_ground_match}), a template $ I_g $ on the ground images is extracted, the size of which is determined by a correlation window $ w_c $. Then, correspondence image subsets of aerial views $I_a$ are also extracted and rectified, using the 3D patch and selected aerial views. The rectified patches are matched against the template $I_g$ using normalized correlation coefficient (NCC) and least-squares matching \citep{gruen1985adaptive,hu2016stable} to refine the aerial-ground matches. 

\subsubsection{Local geometry constraints for ground-synthesized matching}
\label{s:greedy_outlier}

\begin{figure}[h]
    \centering
    \subfigure[Length constraint]{\includegraphics[width=0.35\linewidth]{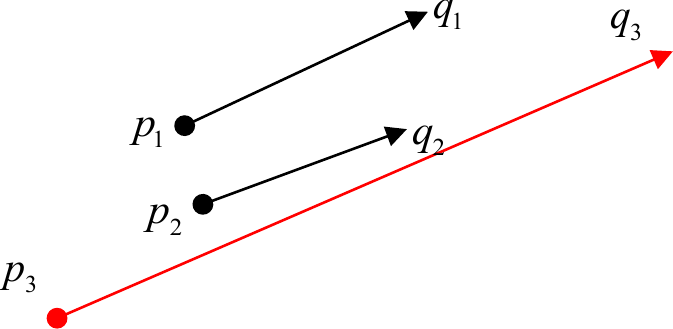}}
    \subfigure[Intersection constraint]{\includegraphics[width=0.3\linewidth]{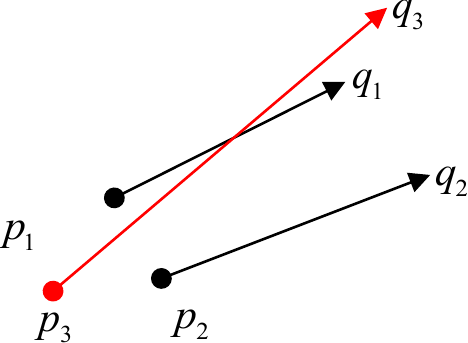}}
    \subfigure[Direction constraint]{\includegraphics[width=0.3\linewidth]{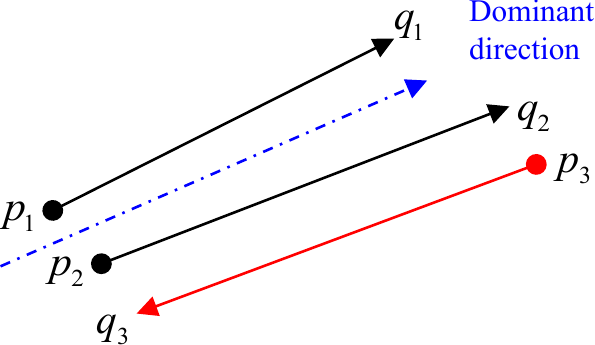}}
    \caption{Constraints for outlier filtering in the matching of ground and synthesized images. The points $ p $ and $ q $ denote the key points in the synthesized and ground images, respectively. Note that $ p $ is placed on the ground image. The red lines indicate matches that violate the constraints.}
    \label{fig:cons_match}
\end{figure}

Due to illumination differences between synthesized and ground images, the SIFT match may contain significantly more outliers after ratio checking, which leads to inferior RANSAC performance. However, because the geometrical differences between the ground and synthesized images are almost negligible, the disparities of correct matches should be small and follow consistent patterns in local regions. Based on these insights, we propose a greedy search algorithm to remove outliers prior to RANSAC. Specifically, from a pair of matched points $ p(x_p,y_p)$ and $ q(x_q,y_q) $, a directed vector can be obtained as $ m=p-q $, which denotes the disparity of the match. If the initial coarse registration is correct, $ m=\boldsymbol{0} $ should be satisfied. However, due to alignment errors and uncompensated distortion, the disparities $ m $ is not exactly zero. But the disparities should at least be consistent with the following three constraints (Figure \ref{fig:cons_match}), which are used sequentially to filter outliers.

\emph{1) Length constraint.} The length of the disparity vector $|m|$ is constrained by an upper limit $\tau_l$, \emph{i.e.} $ |m| < \tau_l $. In practice, $\tau_l$ is chosen as $ 2\% $ of the image extent.

\emph{2) Intersection constraint.} First, we sort the matches by the lengths of $ |m| $ ascendingly. Then, we determine if each segment has an intersection with the $K$-nearest ($K=5$) segments. The segments are indexed using KD-tree. If an intersection exists, the longest segment is marked as outlier.

\emph{3) Direction constraint.} First, we calculate the dominant direction for each segment with respect to the $K$-nearest ($K=5 $) segments. Then, we remove segments that deviate from the dominant direction by an angle $\tau_a$ ($\tau_a=90\degree$ is used), similar to the motion consistency in the work by \cite{jiang2018hierarchical}.

\subsubsection{Propagation of the matches to the aerial images}
\label{s:propagate}

As the meshes are produced from aerial images, the local patches $ p $ should be consistent with all of the aerial images. In theory, directly projecting the 3D point $\boldsymbol{X}$ of the patch $p$ to \textit{suitable} aerial views will obtain correspondences between ground and aerial images. In this paper, three criteria are considered during the selection of \textit{suitable} aerial views, as described below. 

(1) \emph{Containment}, the local patch should locate inside the frustum of the aerial images. This criterion is tested by projecting the four corners of the patch defined by the search window $w_s\delta$ onto all the aerial images. 

(2) \emph{Consistency}, the orientation of the patch $\boldsymbol{n}$ and the direction of aerial image $\textbf{R}^T\boldsymbol{e}_z$ should be consistent, \emph{i.e.} less than a threshold $ \tau_n=90\degree $. This criterion is used because the subset of the aerial images will be rectified locally for the subsequent refinement; if the normal vector of the patch is inconsistent with the aerial view, the rectified image will be blurred due to large perspective deformation. 

(3) \emph{Visibility}, the patch should not be occluded by the mesh itself. For occlusion detection, the optimized bounding volume hierarchy (BVH) of the triangular meshes implemented in Embree \citep{wald2014embree} is used for ray tracing. As BVH structures have almost linear space complexity with regard to the number of triangles, we cache the BVH structure in advance using the meshes that have the finest level of detail. We use OpenSceneGraph \citep{osfield2004open} to load the triangular meshes, which are segmented into small fragments. Then, Geogram \citep{levy2015geogram} is used to automatically clean the fragmented meshes, including welding close vertices and fixing miscellaneous topological defects.

\subsubsection{Matching refinement between aerial and ground images}
\label{s:aerial_ground_match}

\begin{figure}[h]
    \centering
    \includegraphics[width=\linewidth]{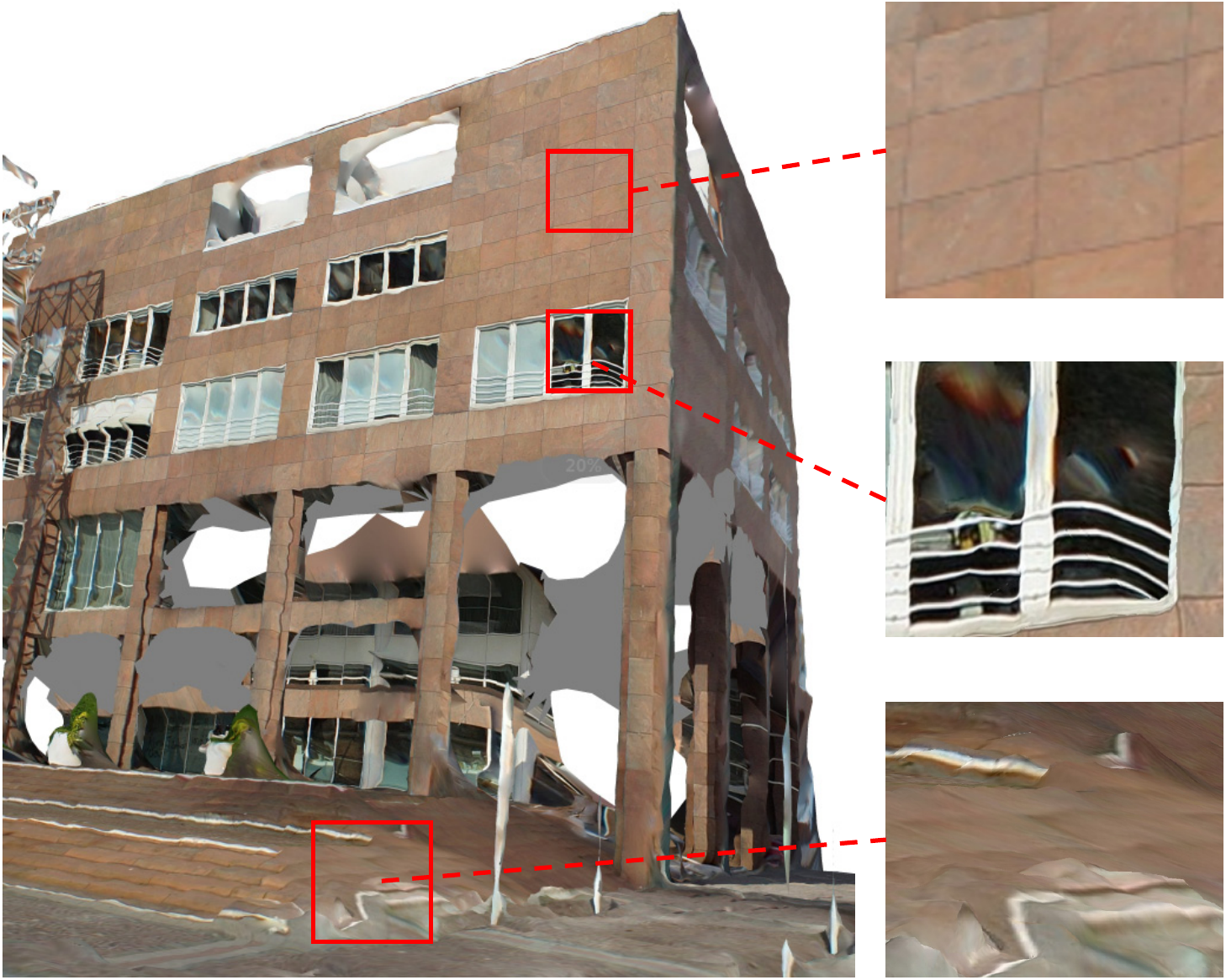}
    \caption{Aspects of the synthesized images that will cause non-negligible errors for aerial-ground matches.}
    \label{fig:issue_synthesized}
\end{figure}

Although the meshes used for rendering are obtained from aerial images, the matches propagated to the aerial images may be inaccurate. The geometry of meshes is noise-laden and the textural information is blended and blurred, as shown in Figure \ref{fig:issue_synthesized}. Therefore, the coordinates of the synthesized images and the corresponding depth value can not be used directly in the combined bundle adjustment. We add an additional step to solve this problem: propagating the matches to aerial images and directly matching the local patches between ground and aerial images. In this way, the matches on the original images will finally be used in the bundle adjustment.

Inspired by the MPGC approach \citep{zhang2005automatic} and our previous view-independent synthesis \citep{hu2015reliable}, we also project all of the patches to the same plane using the homographic transformation $ \textbf{H} $ \citep{hartley2003multiple}: 
\begin{equation}
  \textbf{H}=\textbf{K}_g(\textbf{R}+\boldsymbol{t}\boldsymbol{n}_d^T)\textbf{K}_a^{-1},
  \label{eq:homography}
\end{equation}
where $ \textbf{K} $ is the camera matrix; $ \textbf{R}$ and $ \boldsymbol{t} $ are the relative orientation and translation parameters between the two images, which are deducted from the orientation parameters after coarse registration; $ \boldsymbol{n}_d=\frac{\boldsymbol{n}}{d} $ is the scaled normal vector of the patch, with $\boldsymbol{n}$ the normal vector of the patch and $d$ the distance between patch and aerial view; and the subscripts $ g $ and $ a $ denote the ground and aerial images, respectively. Notably, only the local patches surrounding the initial position are loaded and transformed, rather than the entire images as our previous work \citep{hu2015reliable}.

After rectifying all of the patches, a classic NCC search is used to find the initial match, followed by LSM to further improve the location. The patch extracted from the ground image serves as the template for matching and all of the aerial images are aligned pairwisely. Any match with a correlation smaller than a threshold $ \tau_c $ ($ \tau_c=0.75 $ is used) is pruned before LSM. After LSM, reverse homographic transformation in Equation \ref{eq:ogl_projection} is used to obtain the final coordinates on the aerial images.

\section{Experimental evaluations}

\subsection{Dataset descriptions}

Five datasets (see Table \ref{tab:dataset} and Figure \ref{fig:acquisition}) are used to evaluate the proposed methods, which comprise the ISPRS benchmark dataset collected at Centre of Dortmund and Zeche of Zurich \citep{nex2015isprs} and three datasets collected at the campus of SWJTU. The ground sample distances (GSD) of the images range from $0.2$ to $2.5~cm$. Qualitative and quantitative feature point matching experiments are conducted and compared with existing commercial solutions \citep{acut3d2019context,agisoft2019metashape} and one of the most recent algorithm \citep{revaud2019r2d2}. In addition, to further verify the capability of proposed method, 3D reconstruction results are also presented and compared.

\begin{table}[h]
    \caption{Detailed descriptions of the five datasets used for evaluations.}
    \label{tab:dataset}
	\begin{tabular}{@{}ccccccc@{}}
		\toprule
		\multirow{2}{*}{Dataset} & \multicolumn{2}{c}{Sensor}     & \multicolumn{2}{c}{GSD} ($cm$) & \multicolumn{2}{c}{\#Images} \\
		& Aerial         & Ground        & Aerial            & Ground            & Aerial        & Ground       \\ \midrule
		Centre                   & SONY Nex-7     & SONY Nex-7    & 1.10              & 0.53              & 146           & 204          \\
		Zeche                    & SONY Nex-7     & SONY Nex-7    & 0.56              & 0.28              & 172           & 147          \\
		SWJTU-LIB                & SONY ICLE-5100 & Cannon EOS M6 & 1.69              & 1.06              & 123           & 78           \\
		SWJTU-BLD                & SONY ICLE-5100 & Cannon EOS M6 & 1.93              & 1.33              & 207           & 88           \\
		SWJTU-RES                & SONY ICLE-5100 & DJI spark     & 1.97              & 2.56              & 92            & 192          \\ \bottomrule
	\end{tabular}
\end{table}

The Centre and Zeche datasets are collected by ISPRS in Dortmund and Zurich, respectively. Both the aerial and ground images surrounding a typical building are captured using the same sensor. The other three datasets were all collected in the campus of SWJTU, specifically at the library (SWJTU-LIB), a general building (SWJTU-BLD) and residential areas (SWJTU-RES). Unlike the ISPRS datasets, the aerial images of SWJTU datasets were collected in flights of regular strips and the ground images were captured only for areas of interest. It should be noted that the ground images of SWJTU-RES were not essentially obtained on the ground, which were also captured by a low-cost UAV in a vertical uplift flight. However, because they share similar characteristic of other ground images, we also deem them as ground in this study.

SFM results of both the aerial and ground images are obtained prior to the processing proposed in this paper. In addition, we assume that both image sets are registered roughly; the coarse registration is conducted through either the weak GNSS data obtained in the EXIF header of the images (for Center and Zeche) or three interactively selected tie-points when GNSS data are not available (for the three SWJTU datasets).

\subsection{Evaluation of feature matching}
\subsubsection{Evaluation of feature matching between ground and synthesized images}

\begin{figure}[H]
	\centering
	\includegraphics[width=\linewidth]{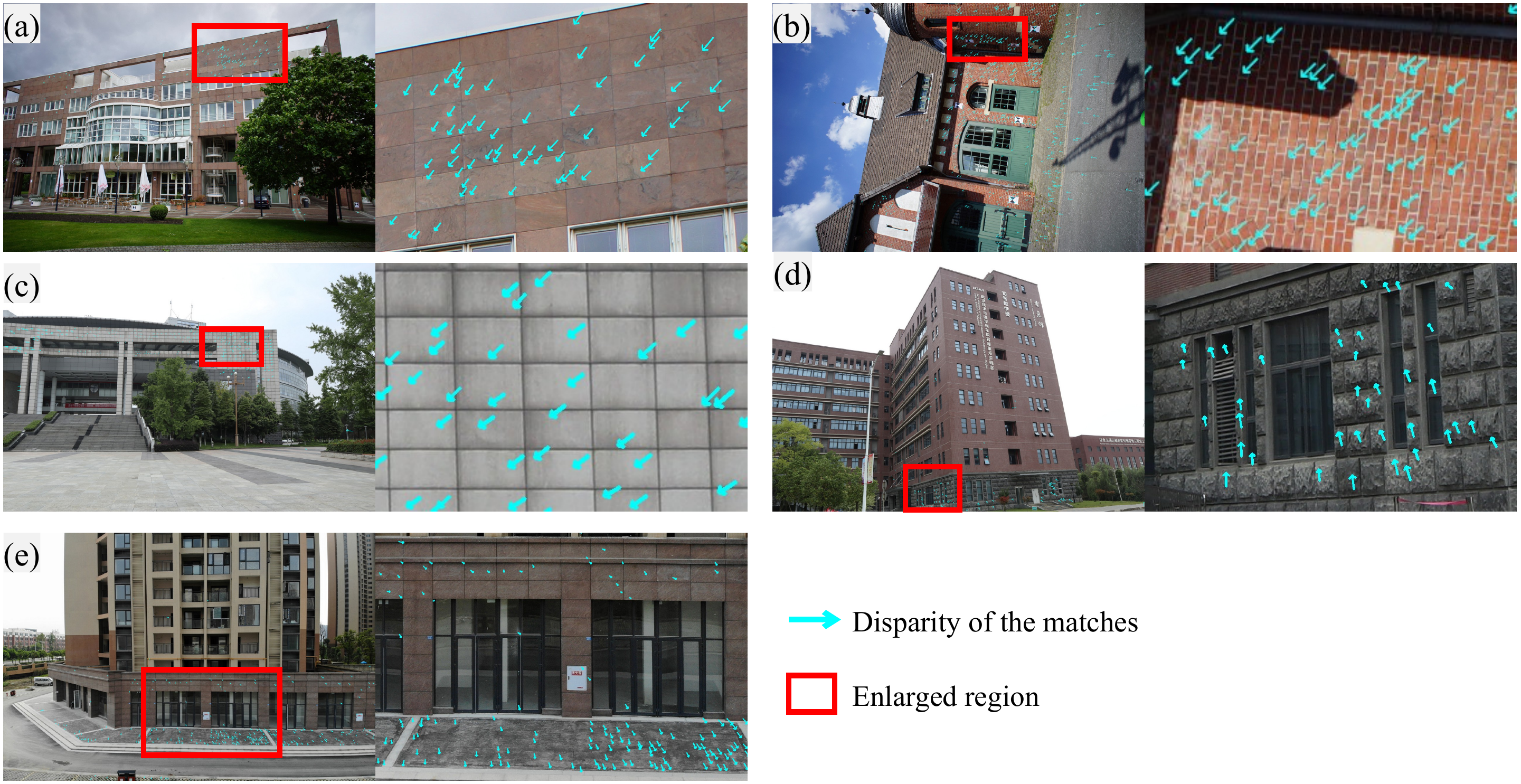}
	\caption{Disparities of the matches between ground and synthesized images drawn on the ground images. (a) to (e) represent results for Centre, Zeche, SWJTU-LIB, SWJTU-BLD and SWJTU-RES, respectively. The arrows are pointing from the coordinates of ground images to those of the synthesized images. Enlarged views indicated by the rectangles are shown on the right part of each subfigure.}
	\label{fig:outlier-disparities}
\end{figure}

Because the synthesized images are obtained using the orientation parameters after coarse registration, the appearances between ground and synthesized images should be similar. In addition, the disparities of the feature matches, \emph{i.e.} the difference of image coordinates, should be small. This is confirmed in Figure \ref{fig:outlier-disparities}, in which the cyan arrows indicate the disparities drawn on the ground images. In fact, the lengths of the disparities can also reflect the accuracies of coarse registration. Another expected characteristic of the distribution of disparities is that they are consistent in local regions, as shown in the enlarged subsets on the right of each subfigure. This is, in fact, the rationale behind the proposed geometric constraints.

\begin{figure}[H]
	\centering
	\includegraphics[width=\linewidth]{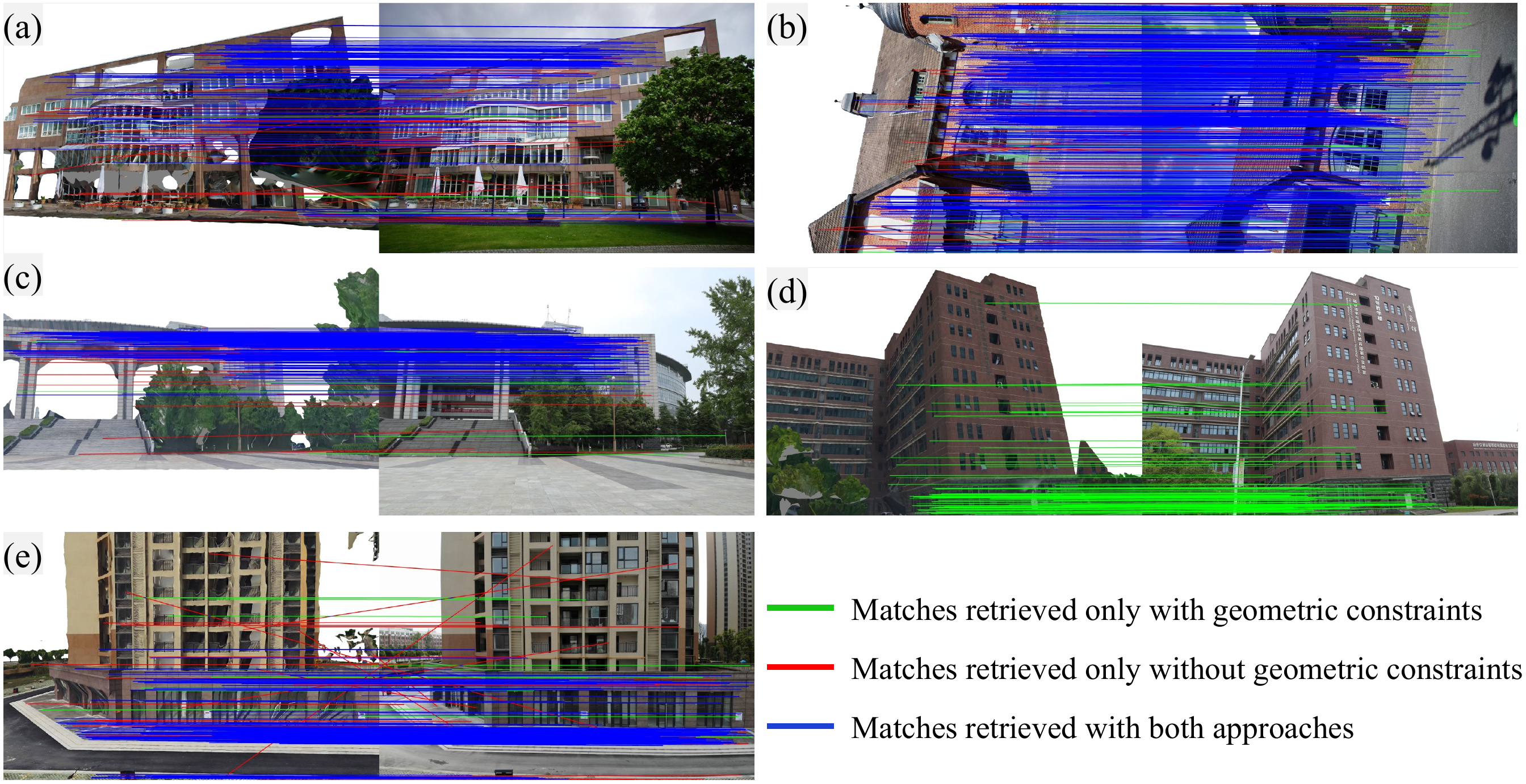}
	\caption{Comparisons of the matches between ground and synthesized images with and without the geometric constraints. (a) to (e) represent results for Centre, Zeche, SWJTU-LIB, SWJTU-BLD and SWJTU-RES, respectively. After ratio checks, the putative matches are categorized into three types: 1) green lines represent matches retrieved only with the geometric constraints; 2) red lines represent matches retrieved only without the geometric constraints; and 3) blue lines represent matches retrieved with both approaches.}	
	\label{fig:outlier-comparisons}
\end{figure}

To evaluate the performances of the proposed geometric constraints in the matching of synthesized and ground images, we compare feature matches with and without the proposed geometric constraints. Figure \ref{fig:outlier-comparisons} shows the matching results for the five datasets. With geometric constraints, the outlier filtering is more stable; we have succeeded in retrieving correct models for all the five datasets, while the SWJTU-BLD is failed without geometric constraint as also demonstrated in Table \ref{tab:outlier}. Notably, even for datasets succeeded without geometric constraints, more outliers are visible, such as Figure \ref{fig:outlier-comparisons}a and e.

\begin{table}[H]
	\centering
	\caption{Comparisons of the outlier filter with and without the proposed geometric constraints in the matching between ground and synthesized images. The values for SIFT are putative matches after ratio checks. The values for the fourth and fifth columns are correct matches after outlier filter.}
	\label{tab:outlier}
	\begin{tabular}{@{}lcccc@{}}
		\toprule
		Dataset   & Image    & \#SIFT & \#Without Constraint & \#With Constraint \\ \midrule
		Centre    & DSC02820 & 1863   & 180      & 152        \\
		Zeche     & DSC04664 & 2685   & 530      & 525        \\
		SWJTU-LIB & DSC01726 & 2152   & 385      & 316        \\
		SWJTU-BLD & IMG1919  & 2111   & 0        & 84         \\
		SWJTU-RES & DJI0137  & 2098   & 266      & 263        \\ \bottomrule
	\end{tabular}
\end{table}

\subsubsection{Evaluation of feature-matching between aerial and ground images}

\begin{figure}[H]
	\centering
	\subfigure[Centre]{\includegraphics[width=0.19\linewidth]{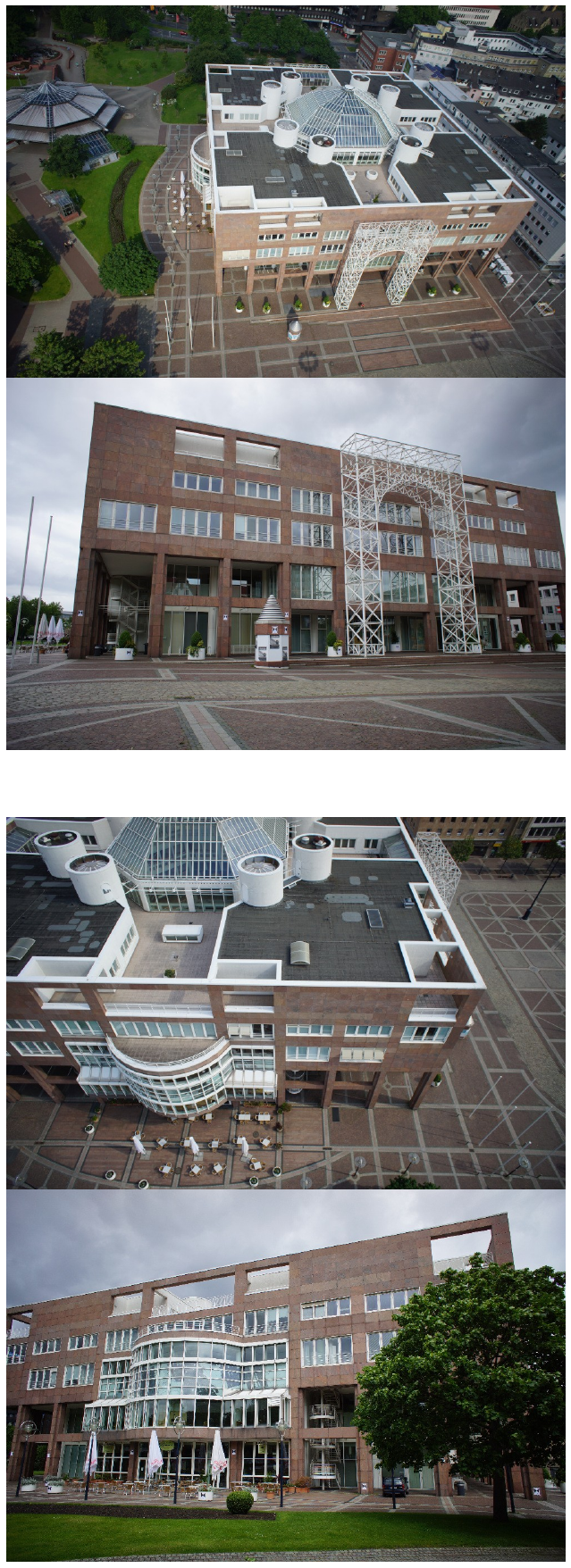}}
	\subfigure[Zeche]{\includegraphics[width=0.19\linewidth]{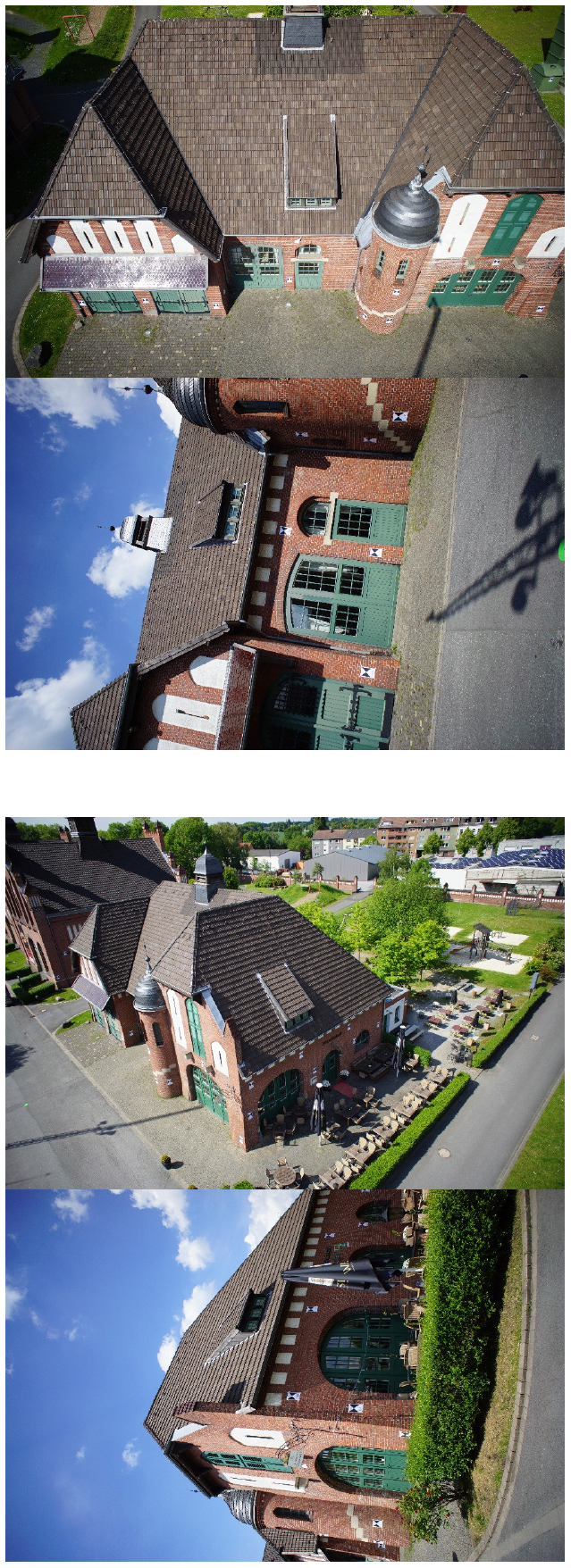}}
	\subfigure[SWJTU-LIB]{\includegraphics[width=0.19\linewidth]{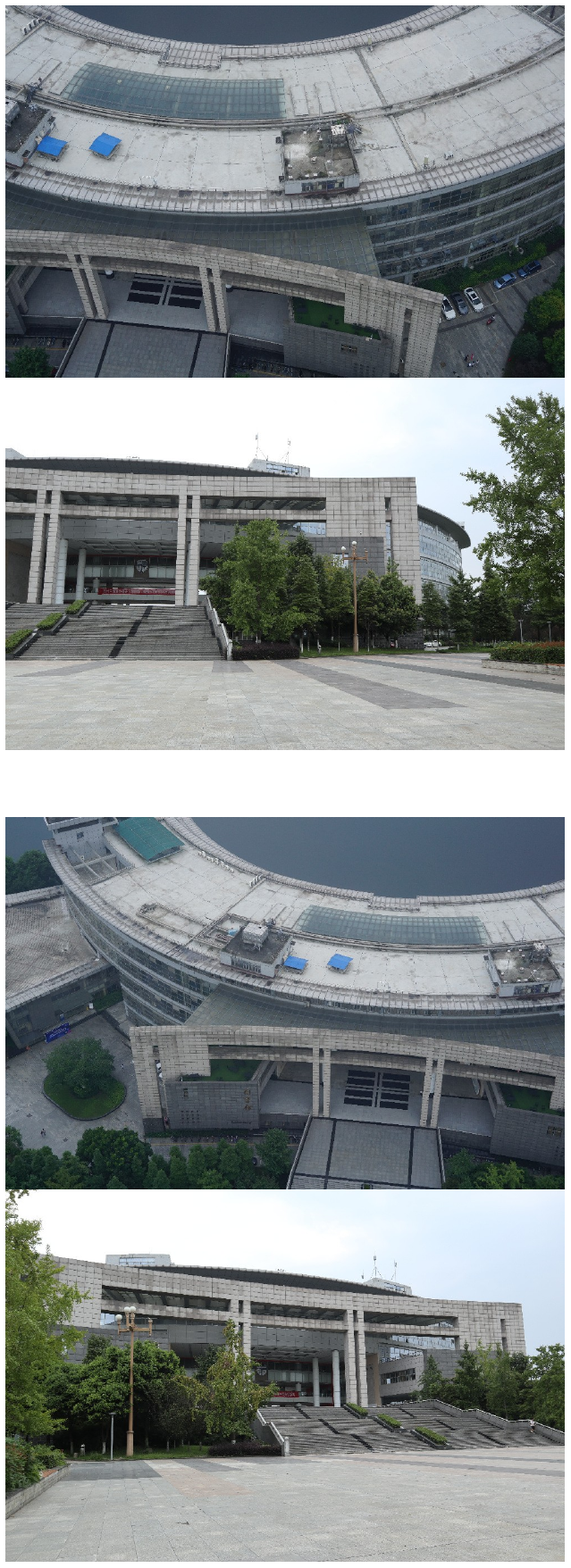}}
	\subfigure[SWJTU-BLD]{\includegraphics[width=0.19\linewidth]{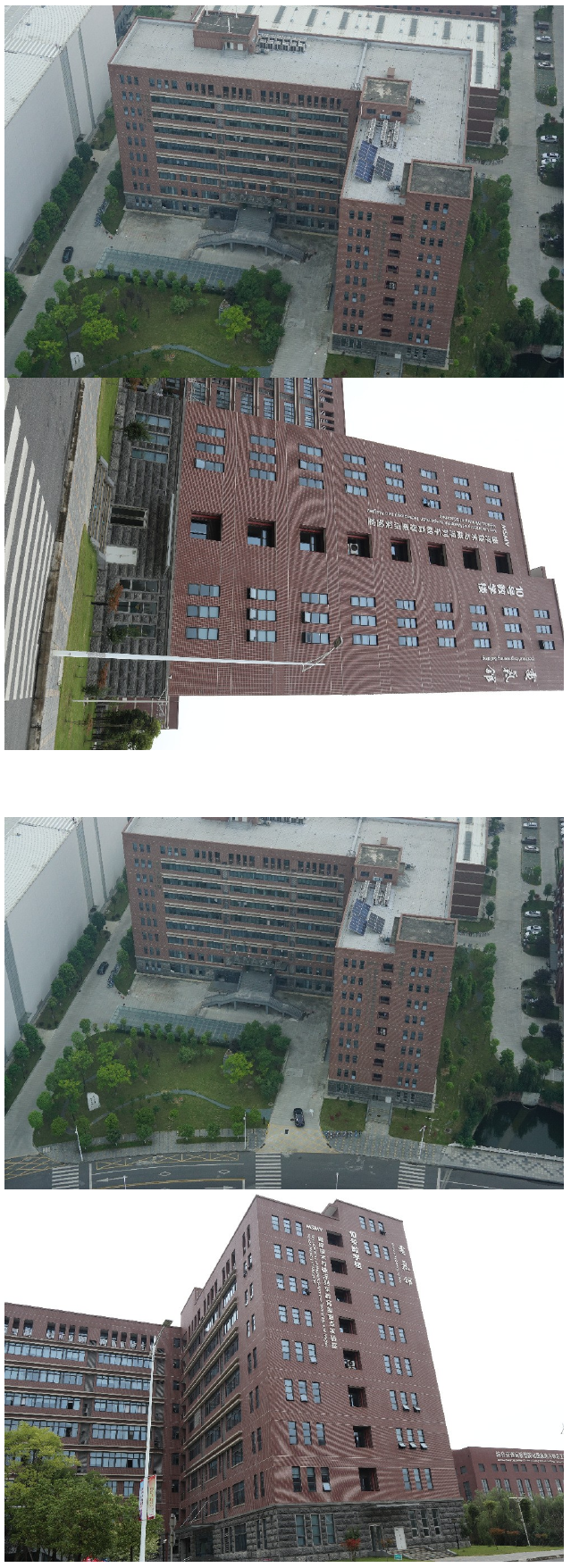}}
	\subfigure[SWJTU-RES]{\includegraphics[width=0.19\linewidth]{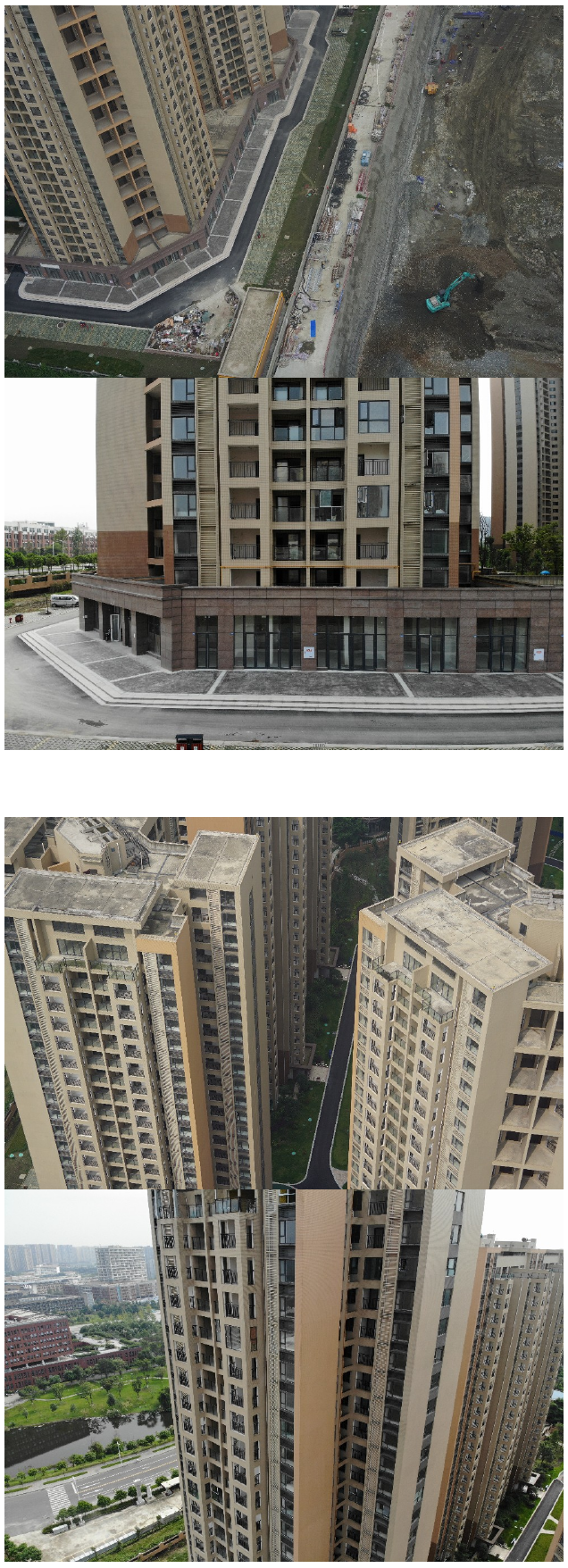}}
	\caption{The selected 10 image pairs from the five test datasets. The odd and even rows show images from aerial and ground sets, respectively.}
	\label{fig:datasets}
\end{figure}

We compare the feature matching results against both SFM pipelines and \textit{ad hoc} features. Five solutions are considered, including the proposed approach, one commercial solution, \emph{i.e.} Agisoft MetaShape \citep{agisoft2019metashape}, two freeware solutions, \emph{i.e.} VisualSFM \citep{wu2011visualsfm} and Colmap \citep{schoenberger2016sfm,schoenberger2016mvs} and a recent feature based on deep learning, \emph{i.e.} R2D2 \citep{revaud2019r2d2}. Ten pairs are selected from the five datasets, with two pairs for each dataset (Figure \ref{fig:datasets}). We prefer pairs with large convergent angles as long as the selected pairs have enough overlaps. As it is possible that the matching results are noise-laden, we manually count the number of correct matches for the ten pairs; the correctness is validated only roughly, such as the same tile of the wall.

Table \ref{tab:match-number} summarizes the results. Notably, the other four solutions often fail in these situations. Thus, although these solutions are quite robust for processing normal scenes or even Internet-scale datasets \citep{schoenberger2016sfm,wu2011visualsfm}, the large perspective deformation between aerial and ground images are still not solved by them. On the contrary, the proposed methods succeeds in all the cases, with convergent angle up to $75\degree$ 

\begin{table}[H]
	\centering
	\caption{Comparisons of the numbers of matches for 10 pairs of images between aerial and ground datasets, in which two pairs are selected for each dataset. The convergent angles for the image pairs are shown in the second row.}
	\label{tab:match-number}
	\begin{tabular}{@{}ccccccccccc@{}}
		\toprule
		Dataset           & \multicolumn{2}{c}{Centre} & \multicolumn{2}{c}{Zeche} & \multicolumn{2}{c}{SWJTU-LIB} & \multicolumn{2}{c}{SWJTU-BLD} & \multicolumn{2}{c}{SWJTU-RES} \\
		Angle ($\degree$) & 50.8         & 61.9        & 40.9        & 51.5        & 54.6          & 61.2          & 59.6          & 70.2          & 34.6          & 75.1          \\ \midrule
		Proposed          & 243          & 114         & 188         & 304         & 91            & 161           & 24            & 5             & 72            & 94            \\
		VisualSFM         & 0            & 12          & 0           & 0           & 12            & 0             & 0             & 0             & 6             & 0             \\
		MetaShape         & 0            & 0           & 0           & 0           & 0             & 0             & 0             & 0             & 0             & 0             \\
		Colmap            & 0            & 17          & 0           & 0           & 29            & 0             & 0             & 0             & 0             & 0             \\
		R2D2              & 17            & 15           & 0           & 0           & 0            & 0             & 0             & 0             & 0             & 0             \\ \bottomrule
	\end{tabular}
\end{table}

\begin{figure}[H]
  \centering
  \subfigure[VisualSFM]{\includegraphics[width=0.3\linewidth]{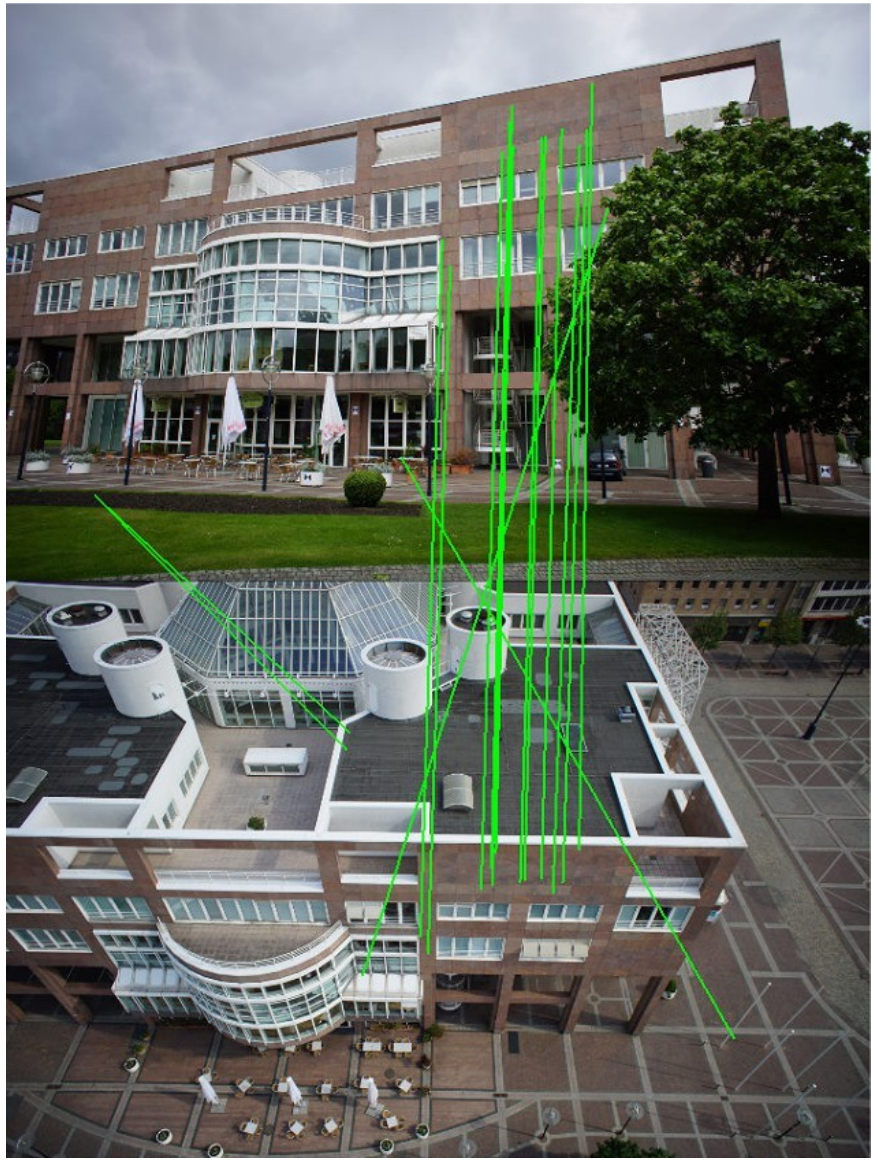}}
  \subfigure[Proposed]{\includegraphics[width=0.3\linewidth]{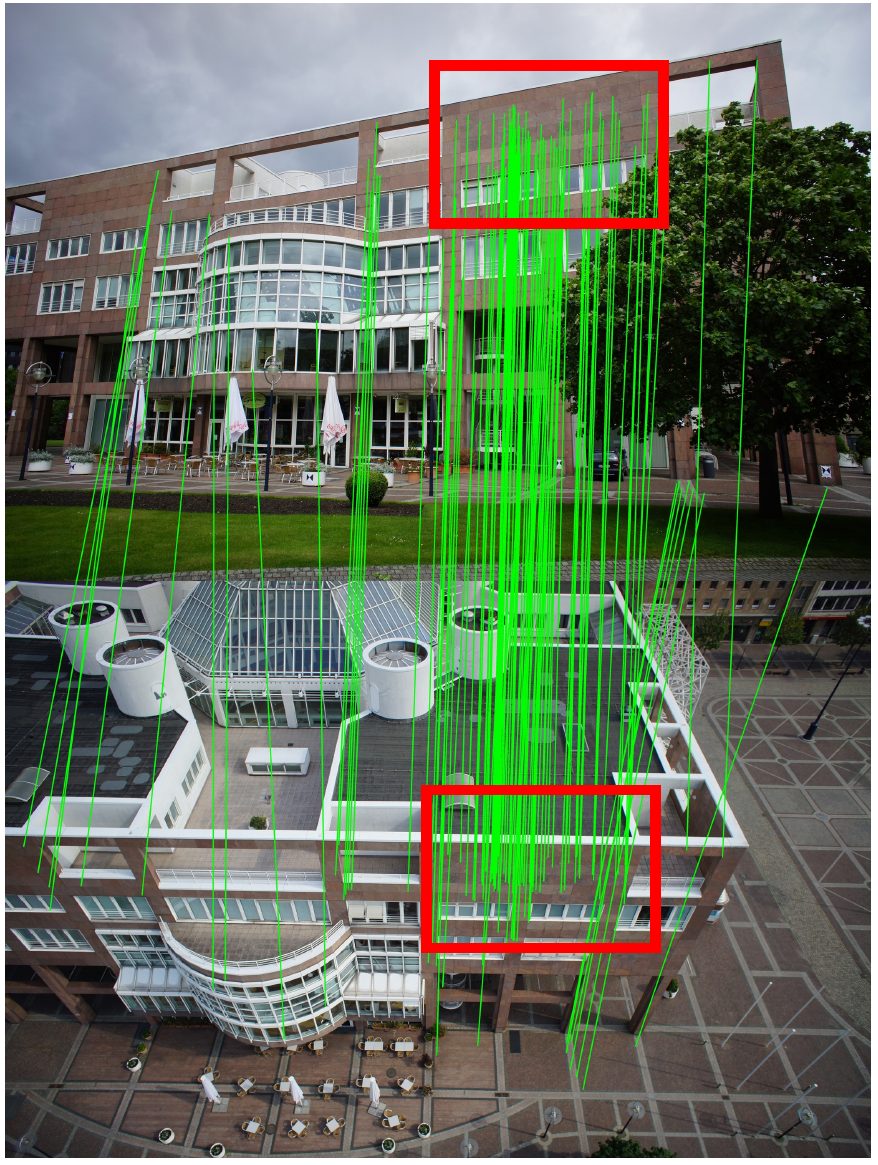}}
  \subfigure[Enlarged]{\includegraphics[width=0.3\linewidth]{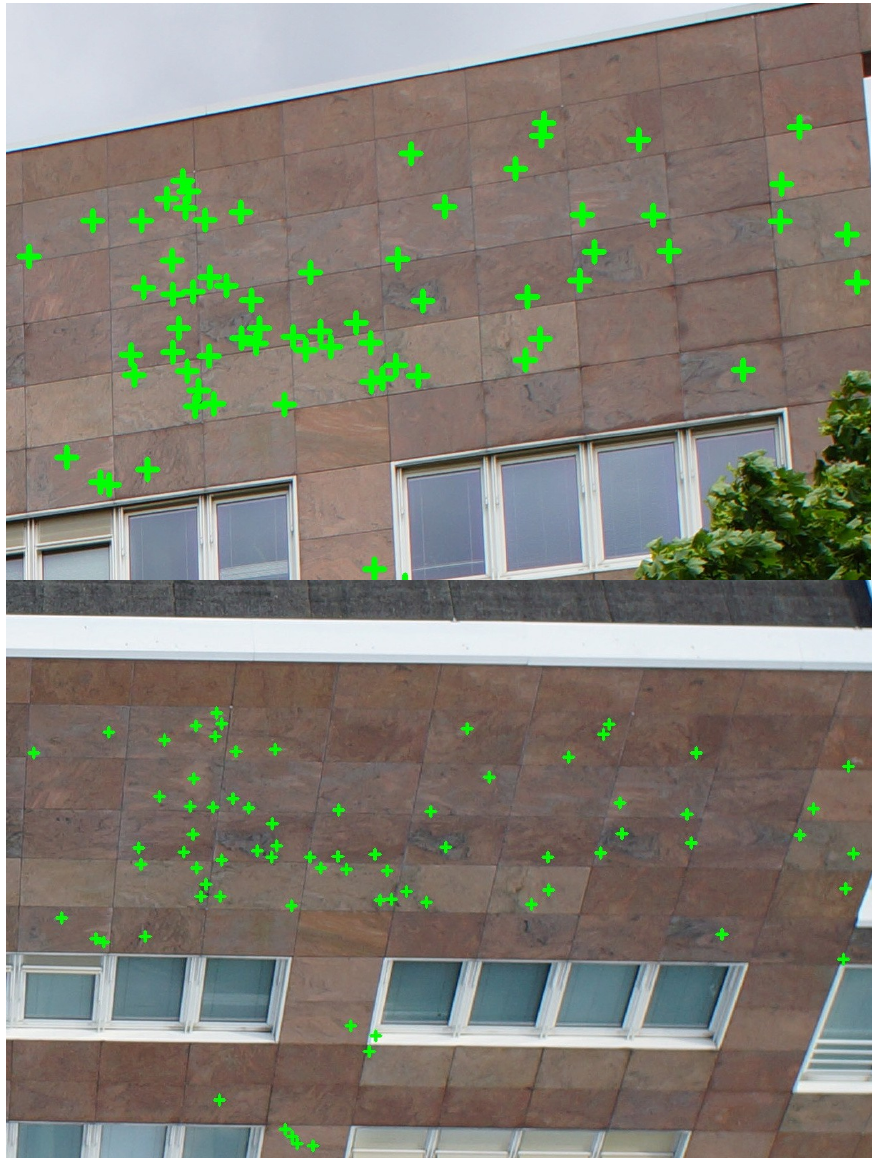}}
  \caption{Aerial-ground matching results for the DSC02820-DSC07379 pair from the Dortmund dataset. The red rectangles denote the enlarged areas. The convergent angle between the two images is $61.9 \degree$.}
  \label{fig:match_centre}
\end{figure}

\begin{figure}[H]
	\centering
	\subfigure[VisualSFM]{\includegraphics[width=0.3\linewidth]{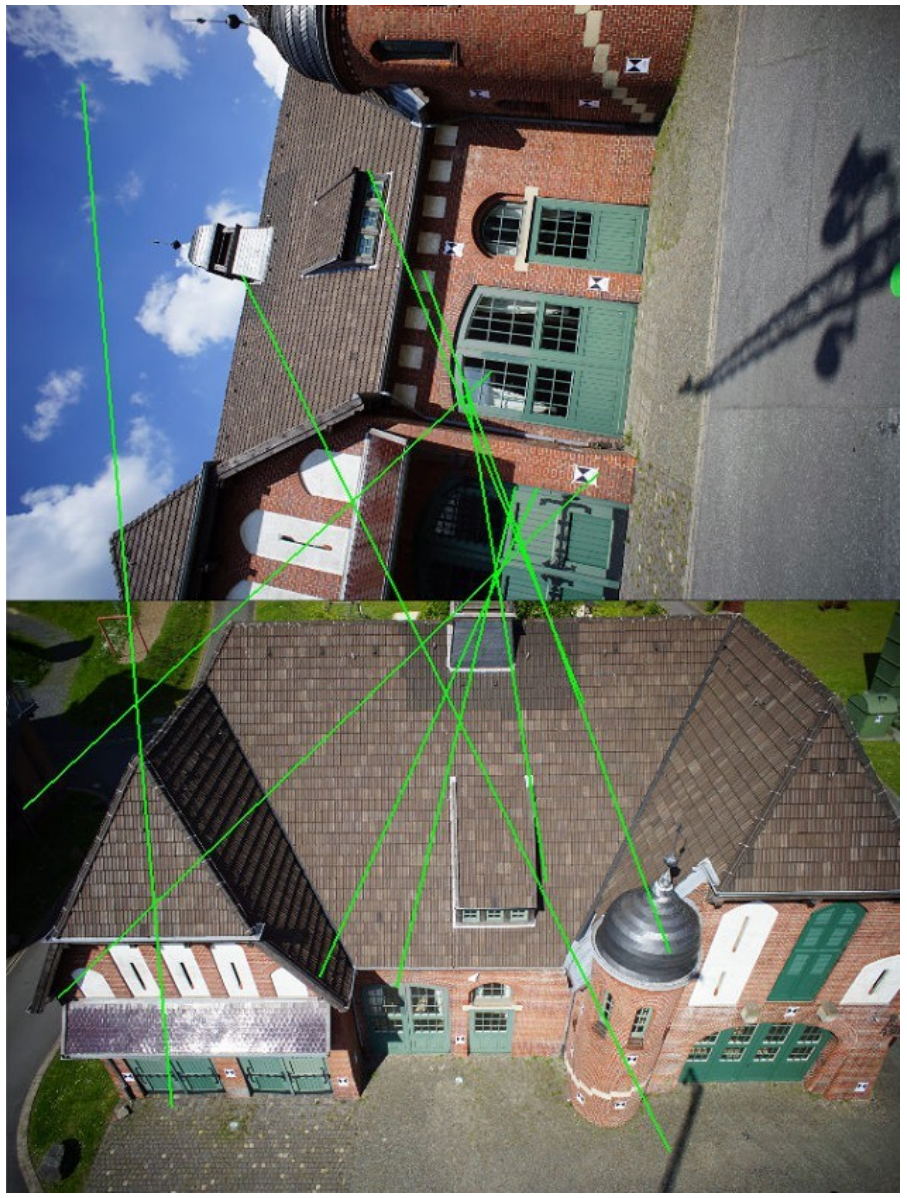}}
	\subfigure[Proposed]{\includegraphics[width=0.3\linewidth]{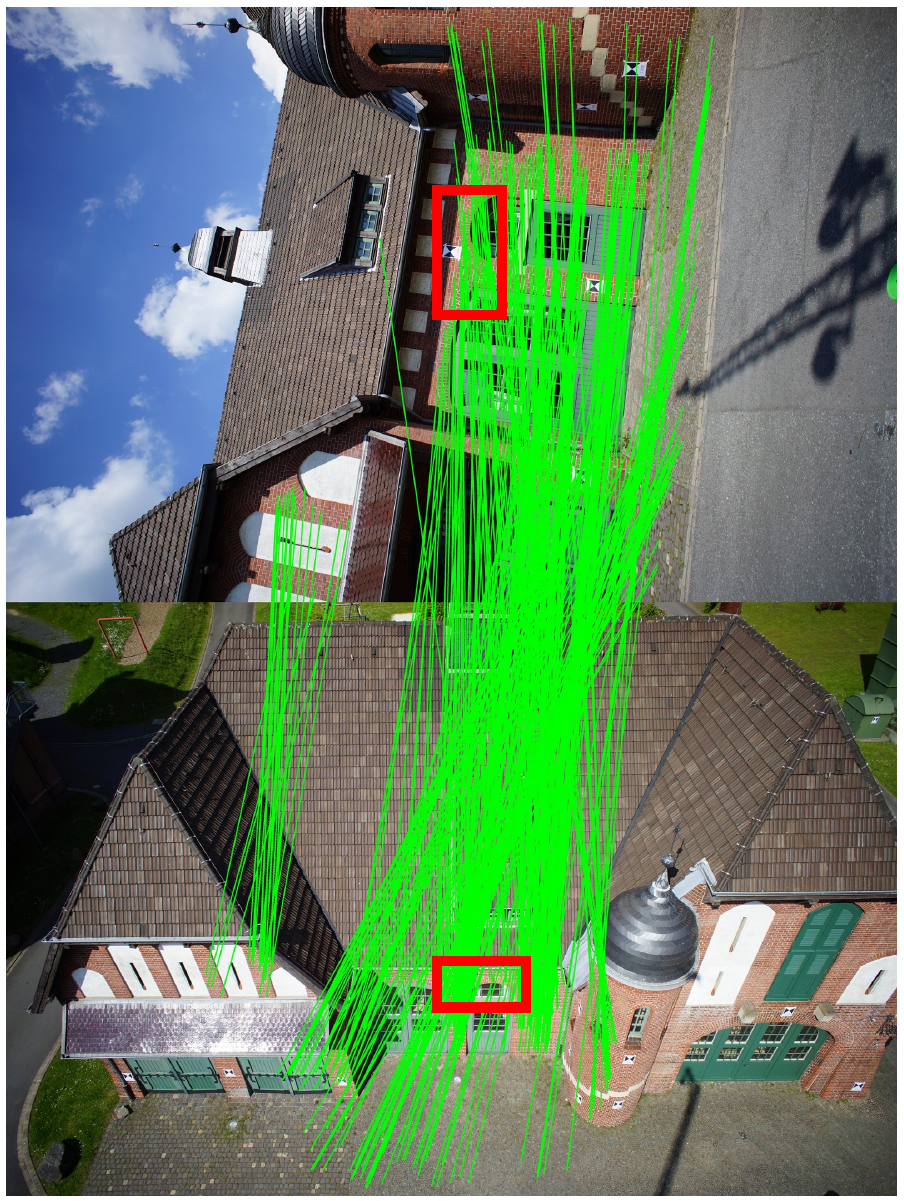}}
	\subfigure[Enlarged]{\includegraphics[width=0.3\linewidth]{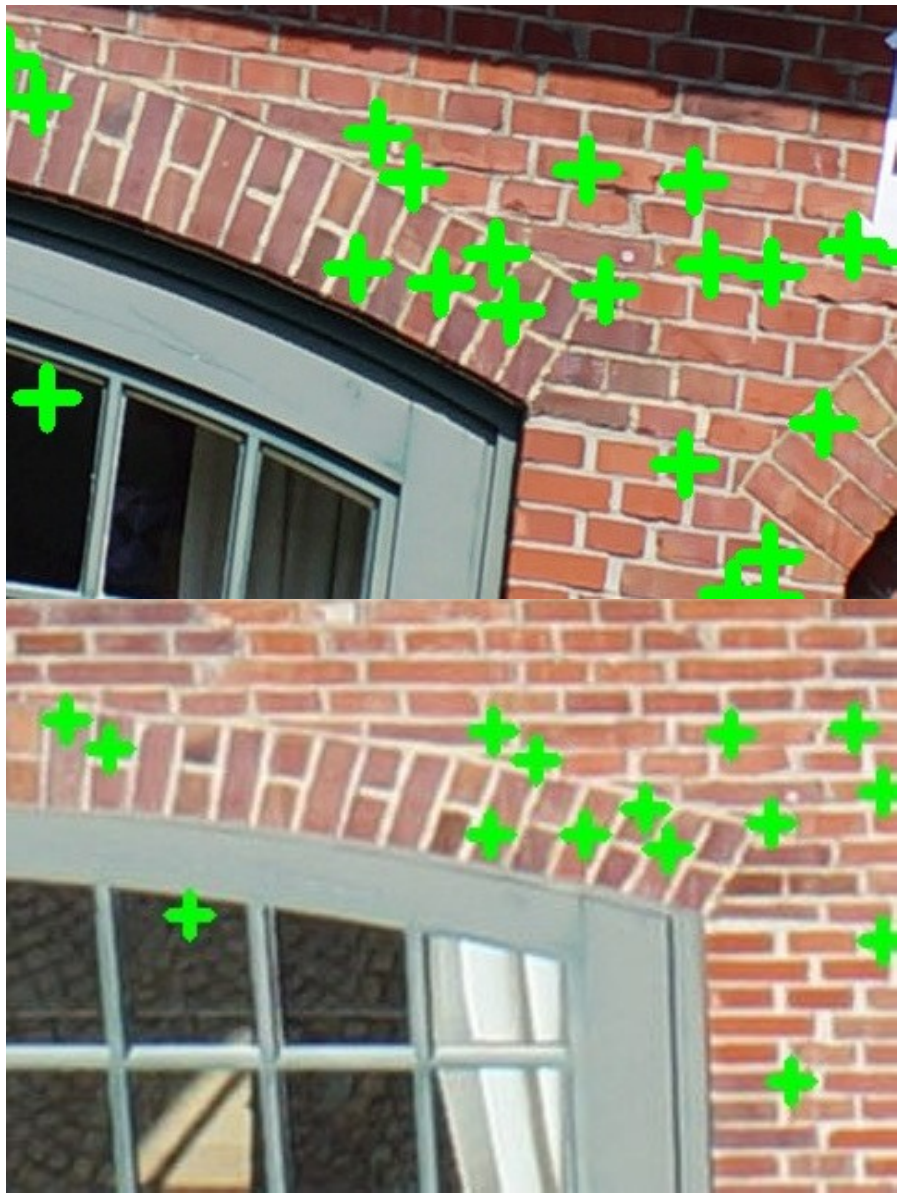}}
	\caption{Aerial-ground matching results for the DSC04664-DSC06239 pair from the Zeche dataset. The red rectangles denote the enlarged areas. The convergent angle between the two images is $51.5\degree$ and the enlarged view for the ground image is rotated $90\degree$ clock-wisely for better visualization.}
	\label{fig:match_zeche}
\end{figure}

\begin{figure}[H]
  \centering
  \subfigure[VisualSFM]{\includegraphics[width=0.3\linewidth]{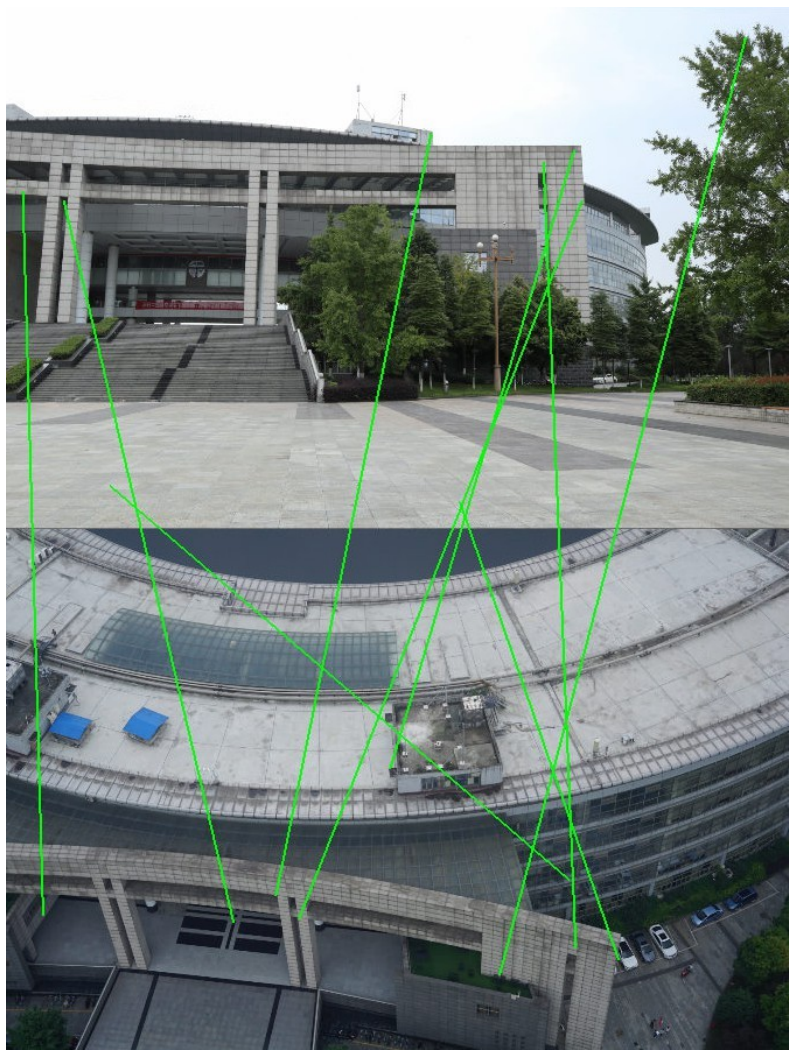}}
  \subfigure[Proposed]{\includegraphics[width=0.3\linewidth]{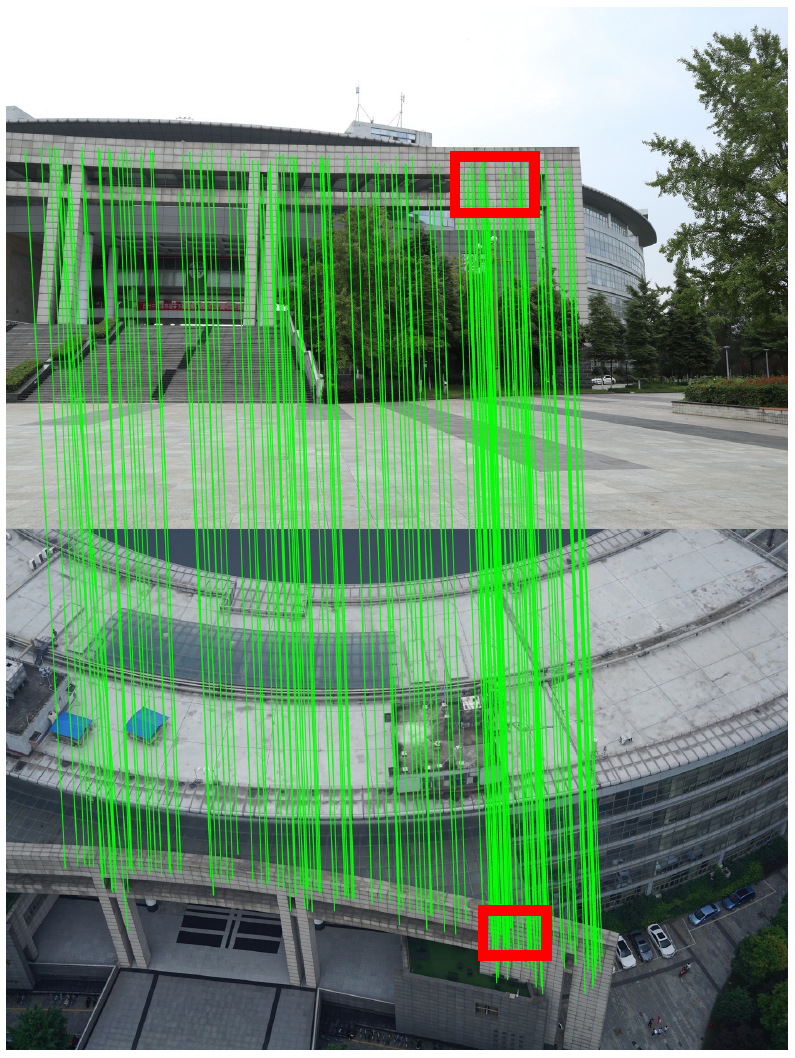}}
  \subfigure[Enlarged]{\includegraphics[width=0.3\linewidth]{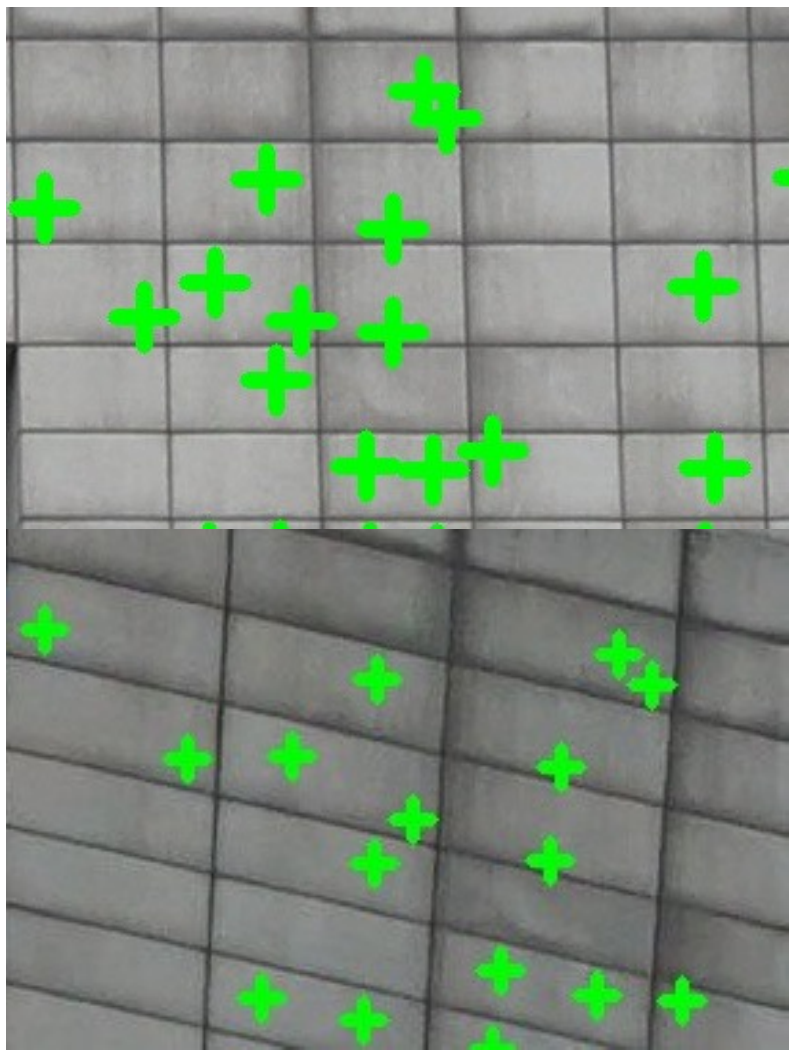}}
  \caption{Aerial-ground matching results for the IMG1726-W0762 pair from the SWJTU-LIB dataset. The red rectangles denote the enlarged areas. The convergent angle between the two images is $61.2\degree$.}
  \label{fig:match_lib}
\end{figure}

\begin{figure}[H]
  \centering
  \subfigure[VisualSFM]{\includegraphics[width=0.3\linewidth]{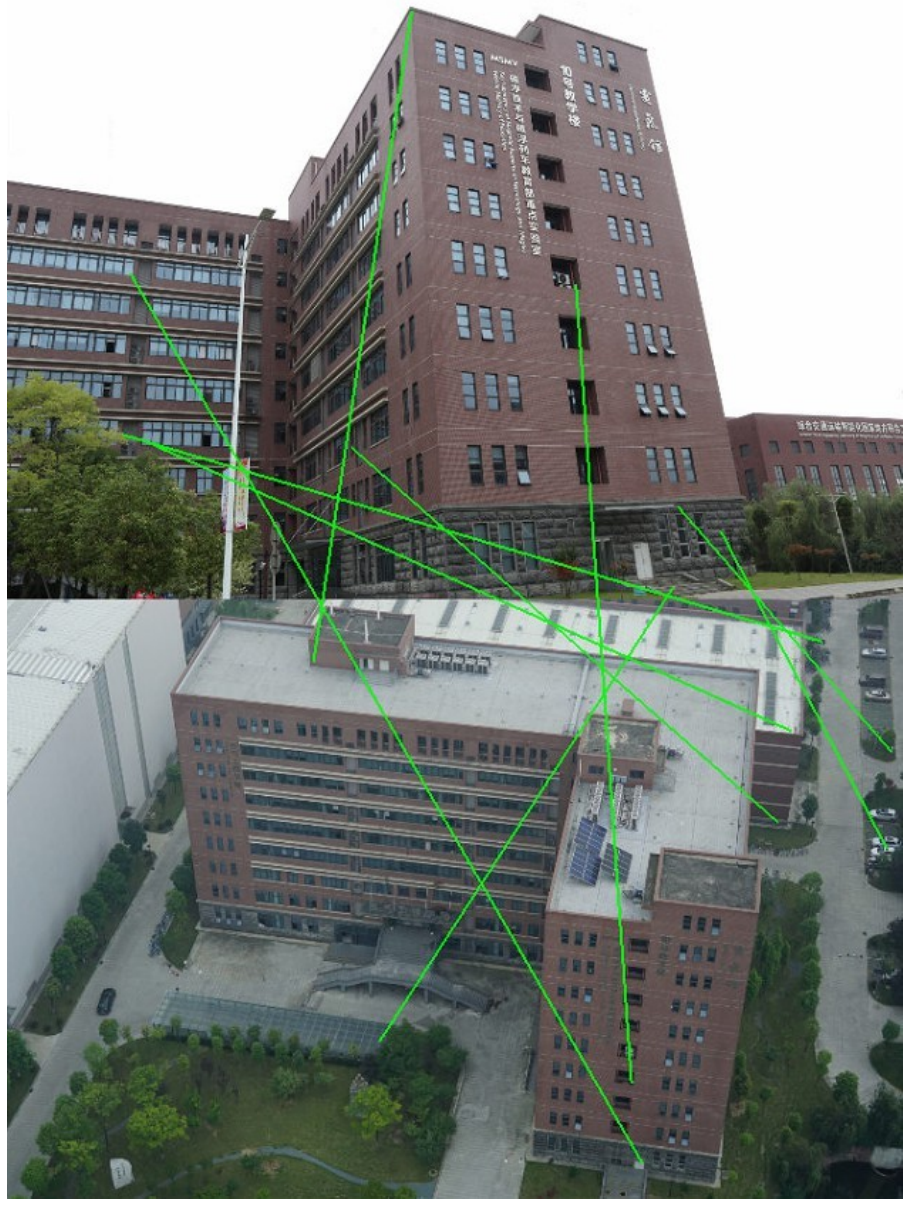}}
  \subfigure[Proposed]{\includegraphics[width=0.3\linewidth]{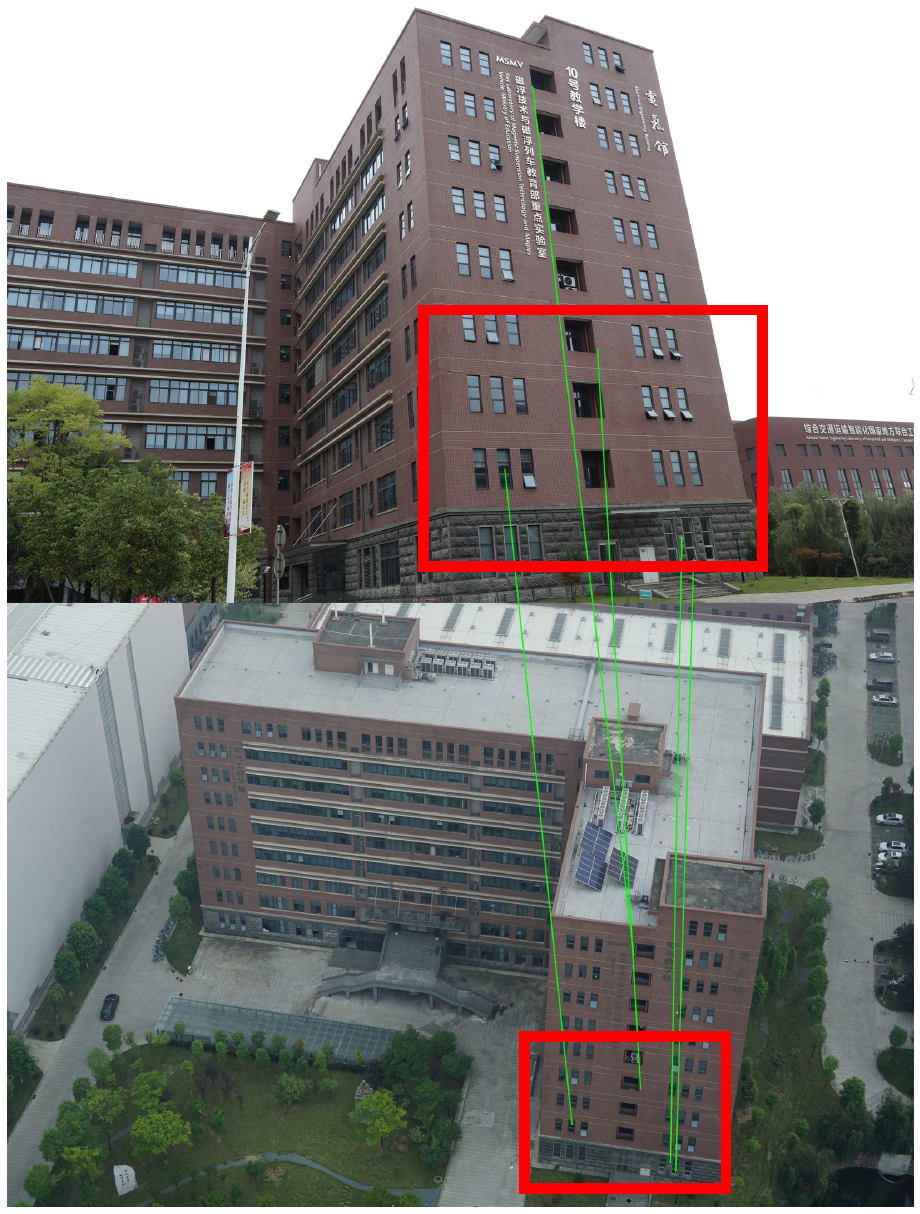}}
  \subfigure[Enlarged]{\includegraphics[width=0.3\linewidth]{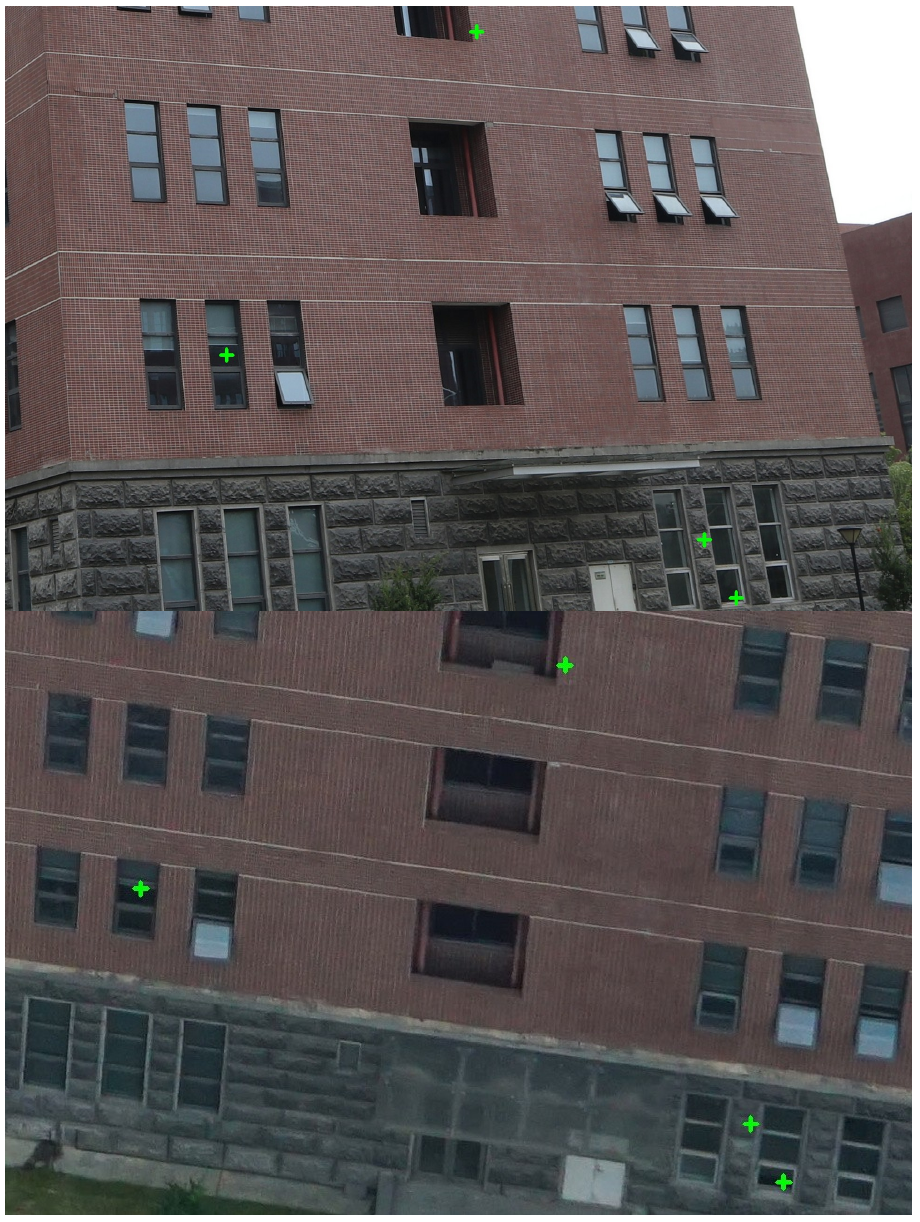}}
  \caption{Aerial-ground matching results for the IMG1919-X0650 pair from the SWJTU-BLD dataset. The red rectangles denote the enlarged areas. The convergent angle between the two images is $70.2\degree$.} 
  \label{fig:match_bld}
\end{figure}

\begin{figure}[H]
  \centering
  \subfigure[VisualSFM]{\includegraphics[width=0.3\linewidth]{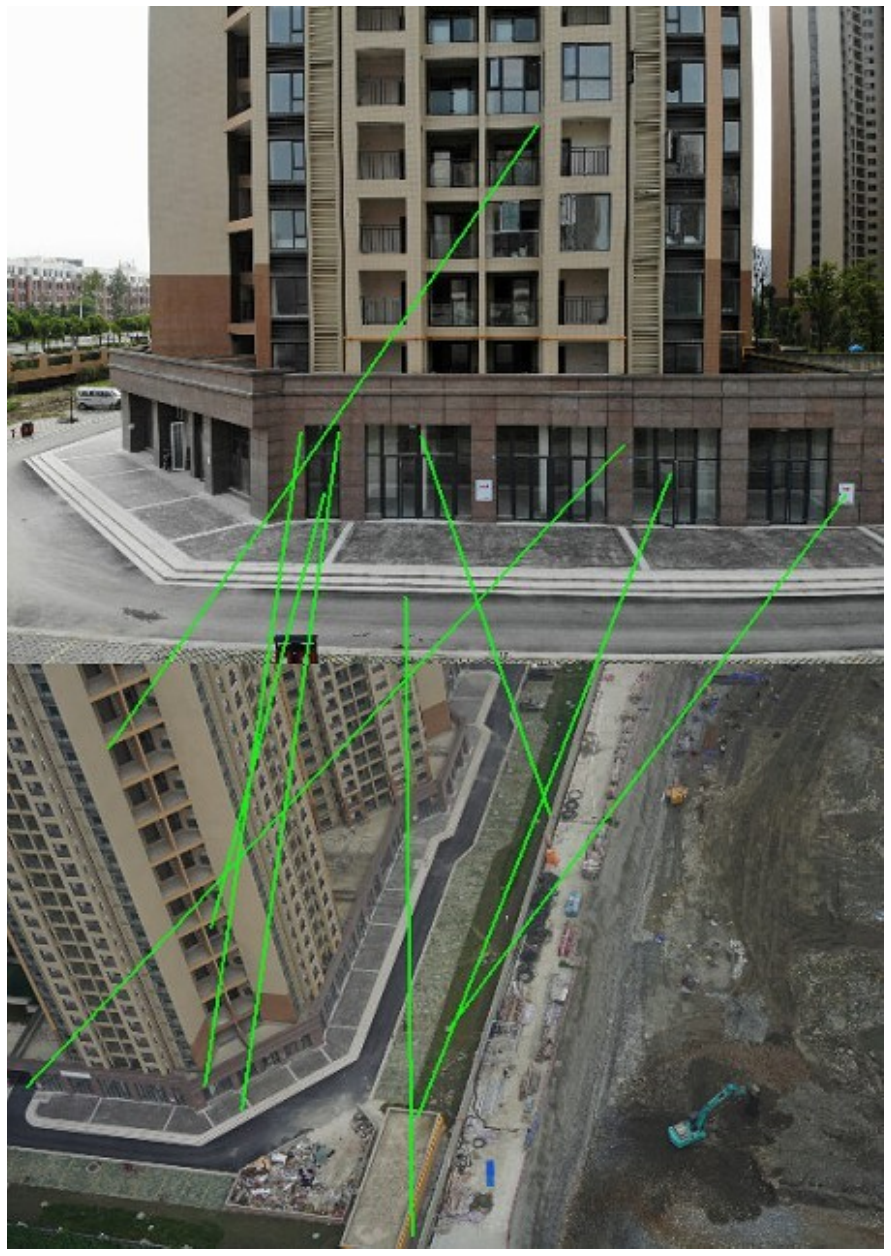}}
  \subfigure[Proposed]{\includegraphics[width=0.3\linewidth]{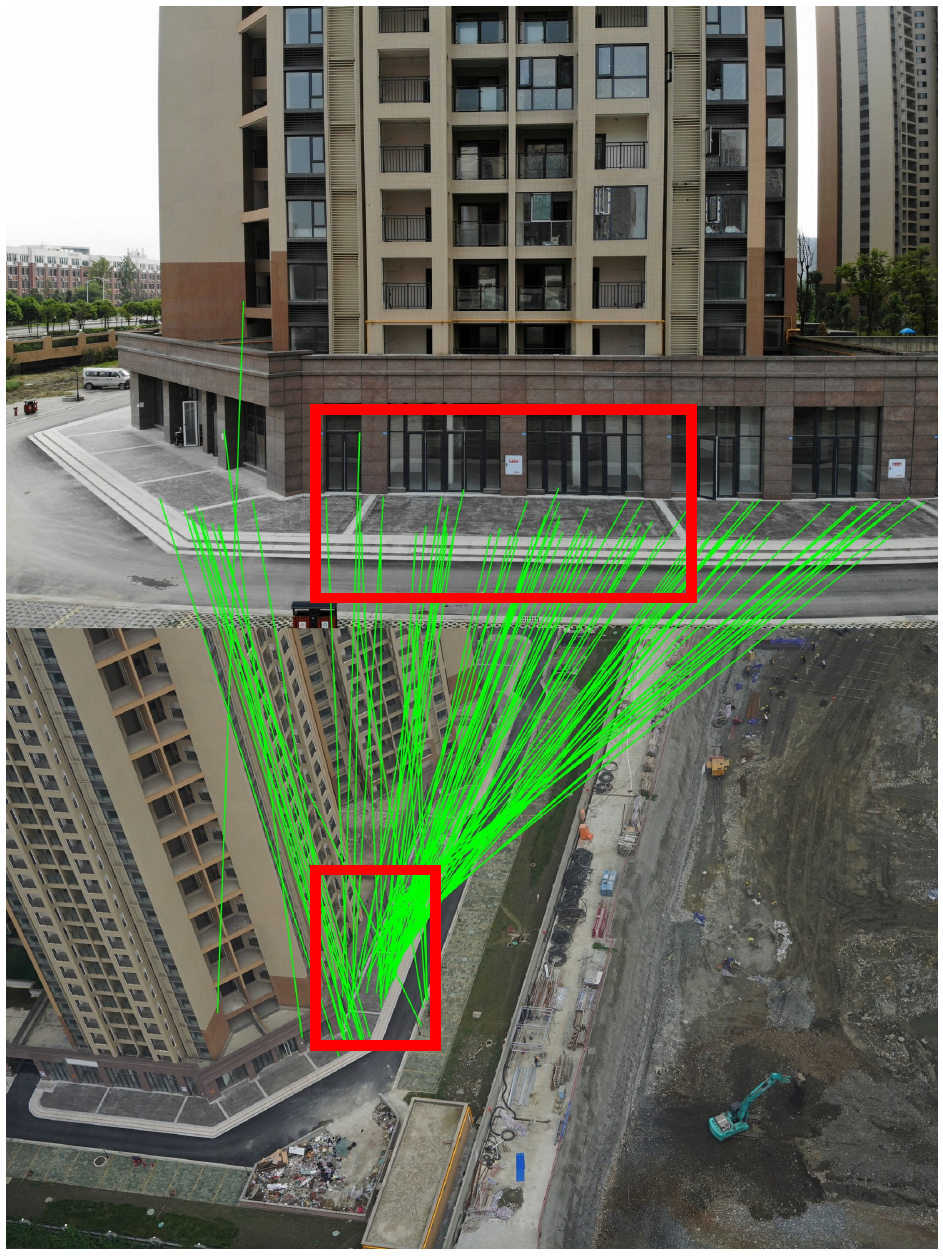}}
  \subfigure[Enlarged]{\includegraphics[width=0.3\linewidth]{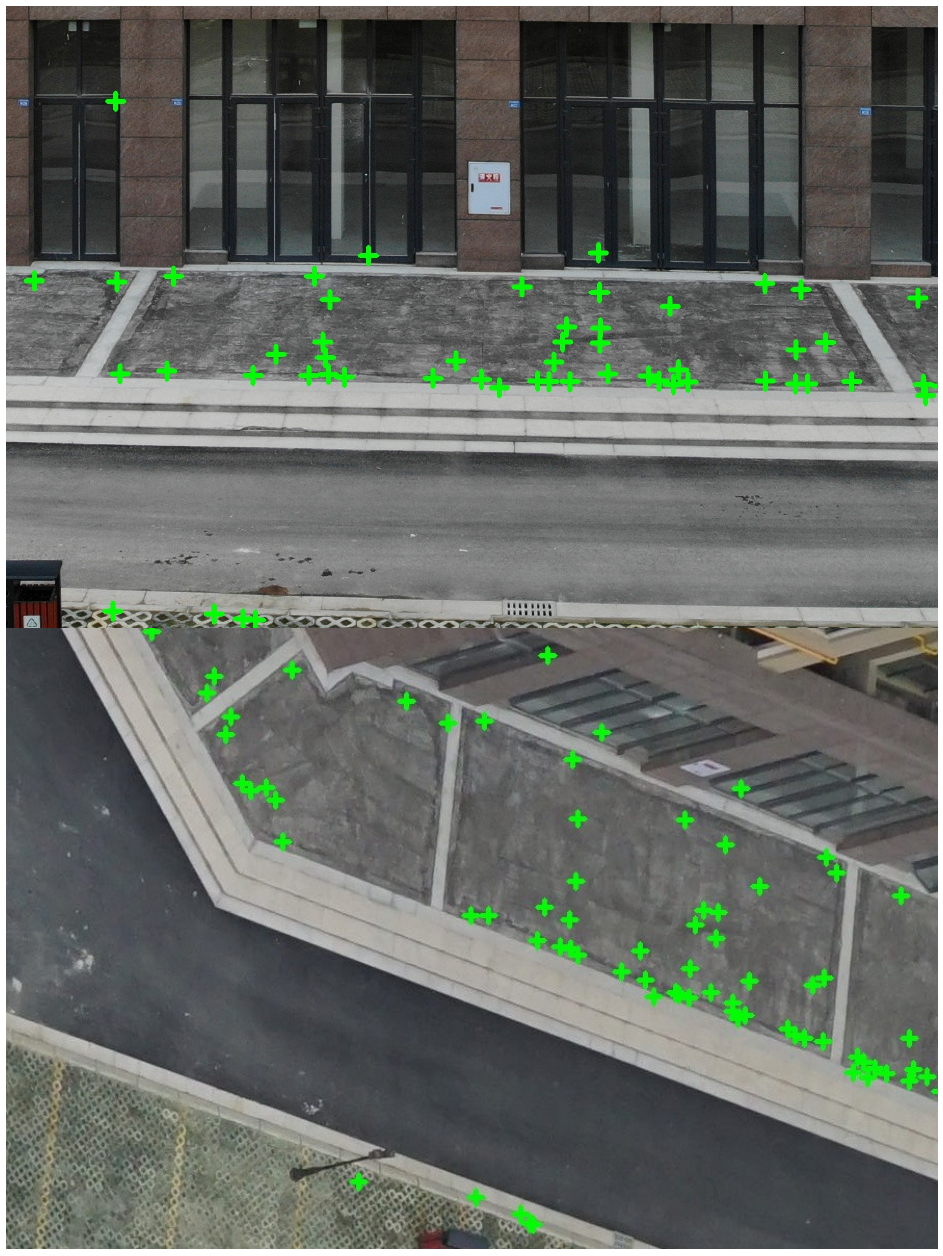}}
  \caption{Aerial-ground matching results for the DJI0312-D0605 pair of the SWJTU-RES dataset. The red rectangles denote the enlarged areas. The convergent angle between the two images is $75.1\degree$ and the enlarged view for the aerial image is rotated $90\degree$ clock-wisely for better visualization.}
  \label{fig:match_res}
\end{figure}

We also select one pair from each dataset and compare the matching results visually against the results afforded by the second-best processing system, VisualSFM, in Figures \ref{fig:match_centre} through \ref{fig:match_res}. During these comparisons, the pair with larger convergent angle in Table \ref{tab:match-number} is chosen. The proposed methods succeeds in obtaining a certain amount of correct matches for all the five pairs; and VisualSFM only manages to obtain some correct matches for the Centre dataset only, with noticeably higher outlier ratio.

We also highlight some interesting and appealing characteristics of the proposed methods in the enlarged regions. 1) \emph{Repeated pattern}, the walls of Centre, Zeche and SWJTU-LIB all demonstrate clear repeated patterns and the proposed approach achieves satisfactory performances in this scenario. 2) \emph{Perspective deformation}, the proposed method is agnostic to perspective deformation as seen in the deformed wall tiles of Centre and SWJTU-LIB; this is because the descriptor searching is only conducted between the ground and synthesized images and template matching and least-squares matching are conducted after rectification guided by the local patch. 3) \emph{Different deformation models}, pairwise rectification based on a common plane \citep{wu2018integration,gao2018ancient} can only alleviate perspective deformation on a single plane, but the proposed method can obtain matches on both the ground and fa\c{c}ades at the same time, as seen in Centre, Zeche and SWJTU-RES. 4) \emph{Glassy objects}, it is arguably that glassy objects are the most challenging cases for image matching, which also causes problem for the proposed approaches; however, we still obtain several correct matches for the SWJTU-BLD dataset; in fact, tens of matches are obtained between ground and synthesized images and five remains after propagating to the aerial view.

\subsubsection{Evaluation of efficiencies of the feature matching}

The time complexity of the feature matching strategy to connect aerial and ground sets of images is $O(n)$, with respect to the number of ground images. On the contrary, simply enumerating all the pairs has time complexity of $O(n^2)$. Considering the large appearant differences between aerial and ground images, the image retrieval technique that achieves time complexity of $O(Kn)$ may not be quite helpful, in which $K$ is a constant for the most similar $K$ images.

However, the runtime of a single pair is absolutely longer due to the additional steps involved. Therefore, we separate the feature matching for a single ground image into three stages: 1) rendering, which consists of loading the mesh models and retrieving all the synthesized images; 2) pairwise matching, which consists of detecting features, descriptor searching and outlier removal and this is a common step involved in almost all feature matching methods; and 3) propagation, which collects visible aerial views, loads the local patches from disks and refines the matches. As shown in Table \ref{tab:runtime}, the costs of additional stages, \emph{e.g.} rendering and propagation, are always \emph{on par} with pairwise matching. The ratios between additional steps and pairwise matching are in the range of (1,2), which indicates that the proposed approach has a linear time complexity, with respect to the number of ground images.

\begin{table}[H]
	\centering
	\caption{Comparisons of different stages of the proposed matching strategy for a ground image. The number of matches are also recorded in the second row and the runtime of last stage is dependent on this number.}
	\label{tab:runtime}
	\begin{tabular}{@{}lcccccccccc@{}}
		\toprule
		Dataset              & \multicolumn{2}{c}{Centre} & \multicolumn{2}{c}{Zeche} & \multicolumn{2}{c}{SWJTU-LIB} & \multicolumn{2}{c}{SWJTU-BLD} & \multicolumn{2}{c}{SWJTU-RES} \\ \midrule
		\#Matches            & 277          & 152         & 349         & 525         & 352            & 316          & 74            & 61            & 151           & 263           \\ \midrule
		Rendering ($s$)      & 2.8          & 8.5         & 2.9         & 3.3         & 6.3            & 5.2          & 2.2           & 5.8           & 1.1           & 0.7           \\
		Pairwise Match ($s$) & 4.0          & 5.6         & 2.5         & 4.2         & 6.2            & 5.7          & 4.2           & 5.9           & 3.6           & 2.7           \\
		Propagation ($s$)    & 1.5          & 3.6         & 5.0         & 8.8         & 11.5           & 4.1          & 1.7           & 0.9           & 1.3           & 1.5           \\ \bottomrule
	\end{tabular}
\end{table}

\subsection{Evaluations of the integrated reconstruction}

We develop an add-on solution for integrated reconstruction, based on ContextCapture \citep{acut3d2019context}. In addition, we also compare three other solutions: the vanilla ContextCapture \citep{acut3d2019context}, MetaShape \citep{agisoft2019metashape} and Colmap \citep{schoenberger2016sfm}. Both sparse and dense reconstructions are evaluated in the following subsections.

\subsubsection{Evaluation of integrated sparse reconstruction}

First, we demonstrate the SFM results by comparing the final numbers of reconstructed images. As some solutions can automatically separate the images into several clusters, only the largest cluster is considered. In addition, we report the number of tie-points that connect aerial and ground images, as these points are the most crucial for the integrated reconstruction. In our experiments, without interactively selected tie-points, most other solutions will not converge to a reasonable results in the SFM procedure. To make a fair comparison, we take about an hour of labor work to add user-input tie-points in the solutions of ContextCapture and MetaShape, for each dataset.

Table \ref{tab:SFM} shows the SFM performances, and it can be seen that the proposed add-on solution for ContextCapture succeeds in all the cases, while the vanilla ContextCapture fails in most of them even with interactively selected tie-points. With user-input tie-points, MetaShape manages to register four out of the five datasets, but the number of tie-points connecting images between aerial and ground sets are fewer than the proposed methods. It is also interesting to see that Colmap succeeds in two datasets even without human interventions using SIFT features; this is probably due to the reliable incremental SFM pipeline \citep{schoenberger2016sfm}. However, we argue that enough tie-points are also important, considering that the proposed approach outperforms other solutions even with a relatively weak SFM solution bundled in ContextCapture. In the Zeche dataset, 31 aerial images are not reconstructed, this is because that the original aerial-only SFM result from ContextCapture does not contain them.
 
\begin{table}[htb]
  \centering
  \caption{Comparisons of different solutions for the five datasets on the sparse reconstruction. The numbers of reconstructed images proportional to the total image numbers are reported in the third and fourth columns. In addition, the numbers of aerial-ground tie-points are presented in the fifth column.}
  \label{tab:SFM}
  \begin{adjustbox}{width=\linewidth}
	\begin{tabular}{@{}clcccc@{}}
		\toprule
		\multirow{2}{*}{Dataset}   & \multicolumn{1}{c}{\multirow{2}{*}{Method}} & \multicolumn{2}{c}{\#Images} & \multirow{2}{*}{\#Aerial-ground tie-points} & \multirow{2}{*}{Status} \\
		& \multicolumn{1}{c}{}                        & Ground               & Aerial              &                                  &                         \\ \midrule
		\multirow{4}{*}{Center}    & Proposed+ContextCapture                     & 203/204              & 146/146             & 23648                            & Succeeded               \\
		& ContextCapture                              & 204/204              & 0/146               & 0                                & Failed                  \\
		& MetaShape                                   & 203/204              & 146/146             & 10428                            & Succeeded               \\
		& Colmap                                      & 168/204              & 0/146               & 0                                & Failed                  \\ \midrule
		\multirow{4}{*}{Zeche}     & Proposed+ContextCapture                     & 172/172              & 116/147             & 38796                            & Succeeded               \\
		& ContextCapture                              & 172/172              & 116/147             & 817                              & Succeeded               \\
		& MetaShape                                   & 172/172              & 147/147             & 23201                            & Succeeded               \\
		& Colmap                                      & 172/172              & 147/147             & 3171                             & Succeeded               \\ \midrule
		\multirow{4}{*}{SWJTU-LIB} & Proposed+ContextCapture                     & 78/78                & 123/123             & 11399                            & Succeeded               \\
		& ContextCapture                              & 78/78                & 123/123             & 20                               & Succeeded               \\
		& MetaShape                                   & 78/78                & 123/123             & 1614                             & Succeeded               \\
		& Colmap                                      & 78/78                & 123/123             & 1374                             & Succeeded               \\ \midrule
		\multirow{4}{*}{SWJTU-BLD} & Proposed+ContextCapture                     & 88/88                & 207/207             & 1706                             & Succeeded               \\
		& ContextCapture                              & 0/88                 & 205/207             & 0                                & Failed                  \\
		& MetaShape                                   & 0/88                 & 207/207             & 0                                & Failed                  \\
		& Colmap                                      & 38/88                & 0/207               & 0                                & Failed                  \\ \midrule
		\multirow{4}{*}{SWJTU-RES} & Proposed+ContextCapture                     & 192/192              & 88/92               & 779                              & Succeeded               \\
		& ContextCapture                              & 192/192              & 0/92                & 0                                & Failed                  \\
		& MetaShape                                   & 192/192              & 91/92               & 323                              & Succeeded               \\
		& Colmap                                      & 0/192                & 16/92               & 0                                & Failed                  \\ \bottomrule
	\end{tabular}
  \end{adjustbox}
\end{table}

To further evaluate the precision and accuracy of the proposed methods, the position uncertainties from the aerial triangulation report and the root-mean-square error (RMSE) of the check points are used. The former (Table \ref{tab:sfm-uncertainty}) metric denotes the internal stability of the SFM results, which is estimated from the covariance matrix  \citep{agarwal2012ceres} of the least-squares solver and taken from the report of ContextCapture. The latter (Table \ref{tab:sfm-accuracy}) denotes performance against external control networks. As different datasets have different accuracies, we also report the results generated using only aerial images as baseline. 

For the uncertainties of image positions (Table \ref{tab:sfm-uncertainty}), almost all the results from aerial-ground integrated approach are better than that of only UAV images, except for SWJTU-BLD; this is probably due to better convergent geometries formed by both aerial and ground images; for SWJTU-BLD, the reason is that the feature matching performances are less robust due to the glassy objects.

For the accuracies of the check points, the results from MetaShape are also compared on the four datasets that MetaShape successfully registered. For each dataset, three or four control points are used in the bundle adjustment, and five to eight check points are used for evaluations. Both control and check points are manually marked at least on three images. Compared to the reference results using UAV images only, both the proposed solution and MetaShape achieved satisfactory results. The proposed solution using ContextCapture as the backend for SFM generally has slightly better horizontal accuracies and MetaShape has better vertical accuracies. 

\begin{table}[H]
	\centering
	\caption{Evaluation of the position uncertainties for each images after bundle adjustment. The values are taken from the report of ContextCapture. For reference, the results from only the aerial images are also demonstrated.}
	\label{tab:sfm-uncertainty}
	\begin{tabular}{@{}lcccccc@{}}
		\toprule
		\multirow{2}{*}{Dataset} & \multicolumn{3}{c}{UAV only ($cm$)} & \multicolumn{3}{c}{Integrated ($cm$)} \\ \cmidrule(l){2-7} 
		& X          & Y          & Z         & X           & Y          & Z          \\ \midrule
		Centre                   & 0.10       & 0.10       & 0.10      & 0.07        & 0.07       & 0.07       \\
		Zeche                    & 0.04       & 0.04       & 0.04      & 0.03        & 0.03       & 0.03       \\
		SWJTU-LIB                & 0.53       & 0.46       & 0.58      & 0.32        & 0.30       & 0.32       \\
		SWJTU-BLD                & 0.58       & 0.56       & 0.45      & 0.71        & 0.77       & 0.59       \\
		SWJTU-RES                & 3.59       & 7.89       & 7.26      & 2.65        & 1.06       & 3.33       \\ \bottomrule
	\end{tabular}
\end{table}

\begin{table}[H]
	\centering
	\caption{Comparisons of the accuracies of check points for the integrated reconstruction. For reference, we also report the results generated using only the aerial images as baseline. The symbol ``-'' indicates missed results due to either lack of check points or failure of the SFM pipeline.}
	\label{tab:sfm-accuracy}
	\begin{tabular}{@{}lccccccccc@{}}
		\toprule
		\multirow{2}{*}{Dataset} & \multicolumn{3}{c}{UAV Only ($cm$)} & \multicolumn{3}{c}{Proposed ($cm$)} & \multicolumn{3}{c}{MetaShape ($cm$)} \\ \cmidrule(l){2-10} 
		& X          & Y         & Z          & X         & Y          & Z          & X          & Y          & Z          \\ \midrule
		Centre                   & -          & -         & -          & 2.6       & 2.0       & 2.2        & 8.3        & 5.9        & 4.8        \\
		Zeche                    & 1.2        & 2.3       & 1.4        & 1.3       & 1.9        & 1.6        & 2.2        & 2.2        & 0.7        \\
		SWJTU-LIB                & 1.0        & 1.1       & 32.1       & 2.4       & 3.3        & 15.5       & 7.8        & 7.5        & 8.8        \\
		SWJTU-BLD                & 1.6        & 1.0       & 4.9        & 3.4       & 9.9        & 12.1       & -          & -          & -          \\
		SWJTU-RES                & 4.7        & 0.9       & 12.7       & 2.7       & 0.7        & 14.5       & 9.7        & 6.6        & 6.5        \\ \bottomrule
	\end{tabular}
\end{table}

\subsubsection{Evaluation of integrated dense reconstruction}

\begin{figure}[htb!]
	\centering
	\subfigure[Dortmund]{\includegraphics[width=0.48\linewidth]{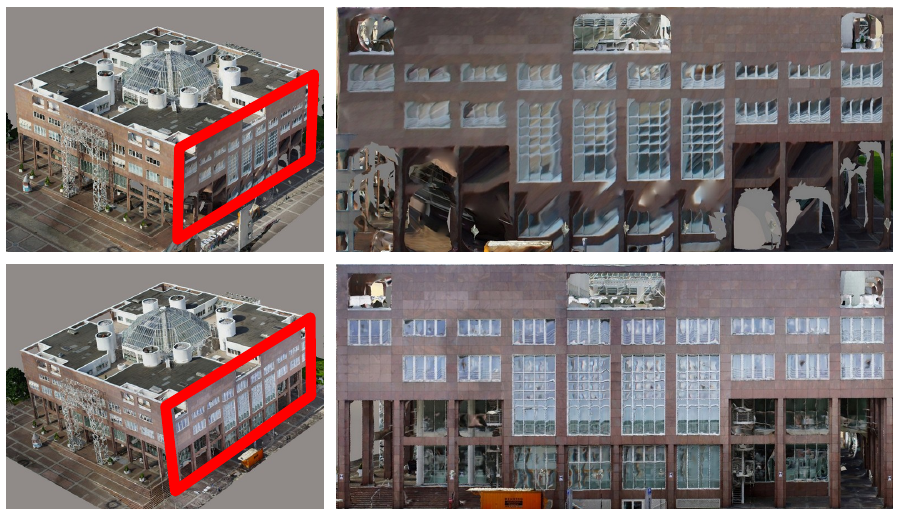}}
	\subfigure[Zeche]{\includegraphics[width=0.48\linewidth]{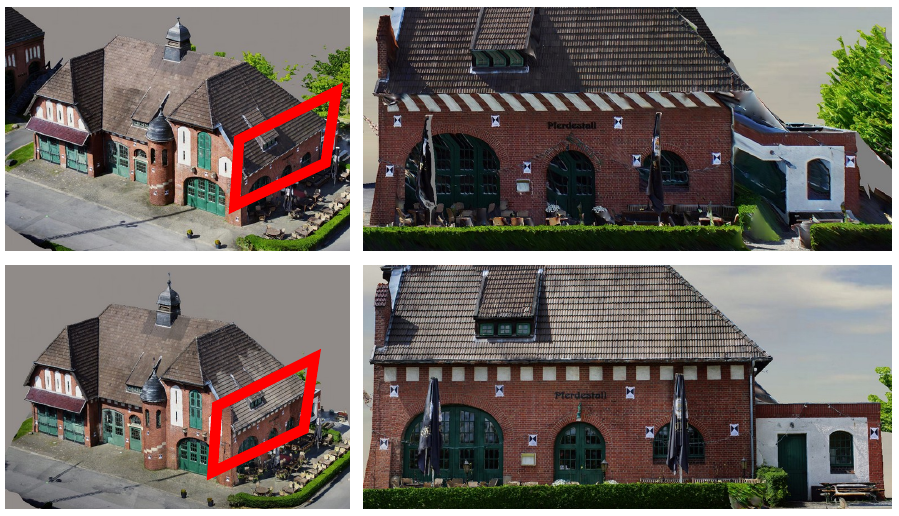}}
	\subfigure[SWJTU-LIB]{\includegraphics[width=0.48\linewidth]{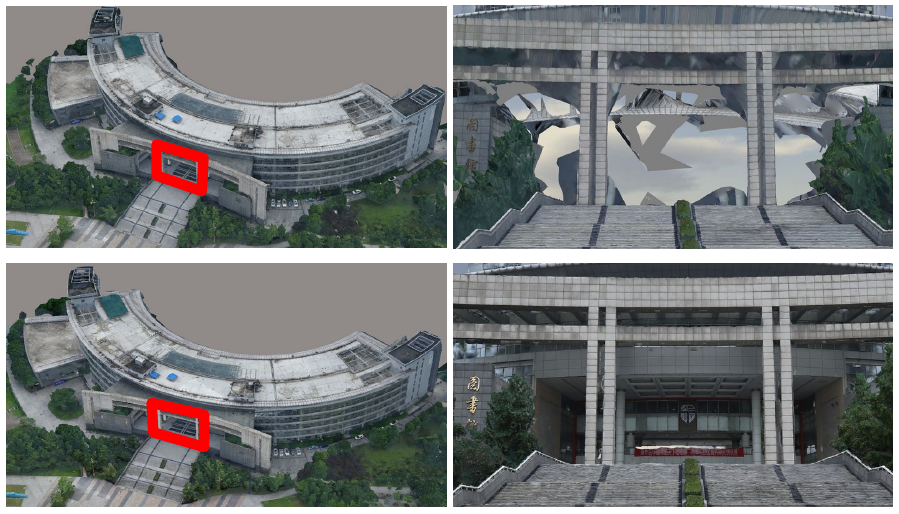}}
	\subfigure[SWJTU-BLD]{\includegraphics[width=0.48\linewidth]{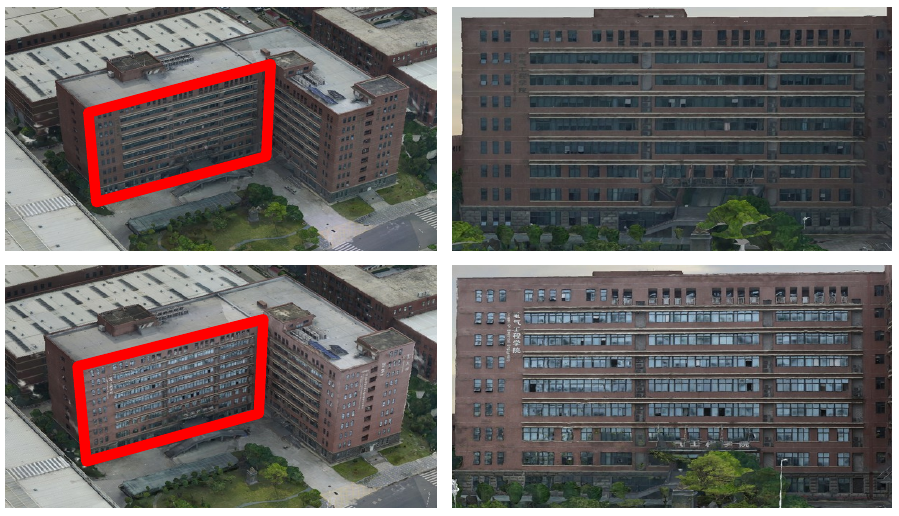}}
	\subfigure[SWJTU-RES]{\includegraphics[width=0.48\linewidth]{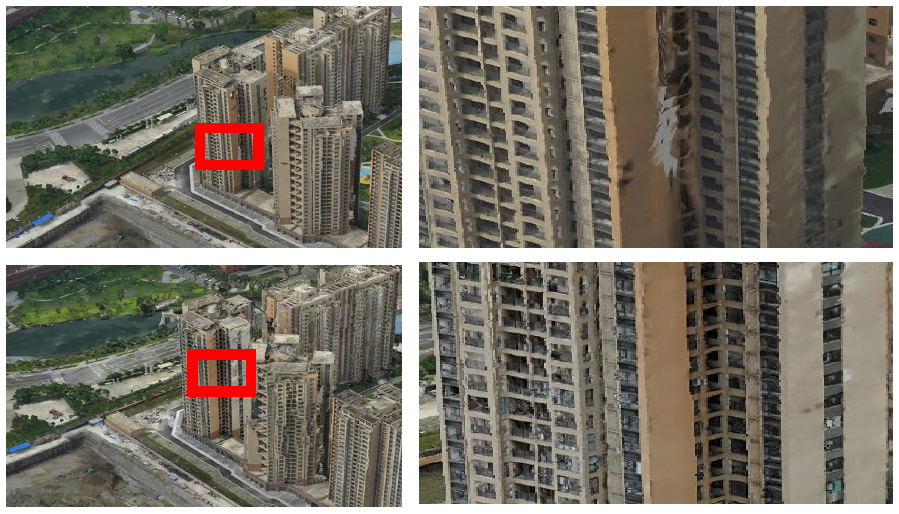}}
	\caption{Comparison of the textured mesh models generated from only aerial images (top row), and those generated from aerial-ground images (bottom row). The right column of each subfigure is an enlargement of the regions highlighted by the rectangles.}
	\label{fig:mesh}
\end{figure}

Figure \ref{fig:mesh} compares the textured mesh models obtained using only aerial images (top row) and integrated solutions (bottom row). We also highlight some parts of the models on the right of each subfigure. Using the integrated solution, the textures on the fa\c{c}ades are clearer, as shown in Figure \ref{fig:mesh}a, c and d. In addition, the reconstructed models are obviously better and more complete, as can be seen in Figure \ref{fig:mesh}c and the small objects in Figure \ref{fig:mesh}d. The quality of texture is also improved, such as the blurred areas under the eaves in Figure \ref{fig:mesh}b.

\subsection{Discussions and limitations}
Based on the above evaluations for feature matching and integrated reconstruction, we summarize some characteristics and limitations of the proposed methods.

\emph{1) Integration with existing SFM and MVS pipeline.}  Although previous solutions \citep{wu2018integration,gao2018ancient} can satisfactorily incorporate aerial and ground images into the same framework, they break existing SFM pipeline and require \textit{ad hoc} bundle adjustment approaches. In fact, the tie-points in the sparse reconstruction are also important for subsequent MVS pipeline, which are used as initial surfaces or constraints, such as the patch-based expanding \citep{furukawa2009accurate}, variational refinement \citep{vu2011high,yu2020fast} or Denaulay triangulation constraints \citep{wu2012integrated}. Instead, the proposed method can be directly used as add-on to existing SFM and MVS pipelines \citep{acut3d2019context}.

\emph{2) Efficiency and accuracy.} The proposed pipeline is also fast and accurate. We do not need to enumerate all the pairs between aerial and ground images, which has quadratic time complexity. Instead, feature matching is only required between ground and synthesized images and is propagated to the aerial views, which has linear time complexity. This is important, because if large viewpoint differences exist, we cannot rely on descriptor-based image retrieval to reduce the numbers of matching pairs. In addition, an additional refinement step is adopted to improve the location of aerial-ground matches.

\emph{3) Limitations.} A limitation of the proposed approach is shared by previous works \citep{wu2018integration,gao2018ancient}, namely that dense reconstruction is required prior to the SFM pipeline. Although our method also requires an additional step, \textit{i.e.} texture mapping, all the above steps are generally bundled in an unified MVS pipeline. In addition, only regions of interest need to be retouched \citep{acut3d2019context} and the runtime overhead may be ignored. Nonetheless, the quality of the textured mesh models will inevitably influence the performance of our approach.

\section{Conclusion}
In this paper, we address the problem of feature matching between aerial and ground images, which currently suffers from severe perspective deformation resulting from viewpoint differences. We elegantly solve the problem by leveraging textured mesh models, which are rendered to the virtual cameras of the ground images. In addition, robust geometric constraints and patch-based matching refinement are used to improve the robustness and quality of the matches. The proposed method is featuring four appealing characteristics: 1) simplicity, the proposed method can be used as add-on solution to existing SFM and MVS pipelines, which simplifies the integration; 2) efficiency, the proposed strategy has linear time complexity rather than quadratic for pairwise rectification \citep{wu2018integration,gao2018ancient}; 3) accurate, the matches are refined locally between aerial and ground images; and 4) robust, the proposed approach is agnostic to the convergent angle between aerial and ground images. Future works may be devoted to further exploiting the possibility of integrating light detection and ranging (LiDAR) point clouds and panoramas collected by the ground mobile-mapping systems into aerial datasets. Code and the SWJTU datasets have been made publicly available at  \url{https://vrlab.org.cn/~hanhu/projects/meshmatch}.

\section*{Acknowledgments}
The authors gratefully acknowledge the provision of the datasets by ISPRS and EuroSDR, which were released in conjunction with the ISPRS Scientific Initiatives 2014 and 2015, led by ISPRS ICWG I/Vb. In addition, this work was supported by the National Natural Science Foundation of China (Projects No.: 41631174, 41871291, 41871314).

\bibliographystyle{model2-names}
\bibliography{MeshMatch}

\end{document}